\documentclass[journal]{IEEEtran}

\usepackage[utf8]{inputenc}
\usepackage[T1]{fontenc}
\usepackage{graphicx}
\usepackage{amsmath,amssymb}
\usepackage{booktabs}
\usepackage{multirow}
\usepackage{dcolumn}
\usepackage{xcolor}
\definecolor{darkgreen}{RGB}{0,128,0}
\usepackage{float}
\usepackage{url}
\usepackage[hidelinks,bookmarksopen=true,bookmarksnumbered=true]{hyperref}

\newcolumntype{.}{D{.}{.}{-1}}
\newcolumntype{d}[1]{D{.}{.}{#1}}
\newcolumntype{e}{D{E}{E}{-1}}
\newcolumntype{E}[1]{D{E}{E}{#1}}

\newcommand{\eqnref}[1]{Eq.(\ref{#1})}

\hypersetup{
  pdftitle={The Aerial-D Dataset for Generalized Referring Expression Segmentation on Aerial Photos},
  pdfauthor={Luís Pedro Soares Marnoto, Alexandre Bernardino, Bruno Emanuel da Graça Martins},
  pdfcreator={LaTeX with hyperref}
}

\begin{document}

\title{Generalized Referring Expression Segmentation on Aerial Photos}
\author{Luís Marnoto, Alexandre Bernardino, and Bruno Martins%
  \thanks{L. Marnoto and B. Martins are with INESC-ID and Instituto Superior Técnico, University of Lisbon, Lisbon 1049-001, Portugal (e-mail: luis.marnoto.gaspar.lopes@tecnico.ulisboa.pt; bruno.g.martins@tecnico.ulisboa.pt).}%
  \thanks{A. Bernardino is with Instituto de Sistemas e Robótica and Instituto Superior Técnico, University of Lisbon, Lisbon 1049-001, Portugal (e-mail: alexandre.bernardino@tecnico.ulisboa.pt).}%
}

\maketitle

\begin{abstract}
Referring expression segmentation is a fundamental task in computer vision that integrates natural language understanding with precise visual localization of target regions. Considering aerial imagery (e.g., modern aerial photos collected through drones, historical photos from aerial archives, high-resolution satellite imagery, etc.) presents unique challenges because spatial resolution varies widely across datasets, the use of color is not consistent, targets often shrink to only a few pixels, and scenes contain very high object densities and objects with partial occlusions. This work presents Aerial-D, a new large-scale referring expression segmentation dataset for aerial imagery, comprising 37,288 images with 1,522,523 referring expressions that cover 259,709 annotated targets, spanning across individual object instances, groups of instances, and semantic regions covering 21 distinct classes that range from vehicles and infrastructure to land coverage types. The dataset was constructed through a fully automatic pipeline that combines systematic rule-based expression generation with a Large Language Model (LLM) enhancement procedure that enriched both the linguistic variety and the focus on visual details within the referring expressions. Filters were additionally used to simulate historic imaging conditions for each scene. We adopted the RSRefSeg architecture, featuring a SigLIP2 vision--language encoder and a SAM segmentation decoder, and trained models on Aerial-D together with prior aerial datasets, yielding unified instance and semantic segmentation from text for both modern and historical images. Results show that the combined training achieves competitive performance on contemporary benchmarks, while maintaining strong accuracy under monochrome, sepia, and grainy degradations that appear in archival aerial photography. The dataset, trained models, and complete software pipeline are publicly available at \href{https://luispl77.github.io/aerial-d/}{\texttt{luispl77.github.io/aerial-d}}.

\vspace{0.3cm}

\end{abstract}


\section{Introduction}
\label{sec:intro}

\IEEEPARstart{R}{eferring} expression segmentation~\cite{kazemzadeh2014referit,hu2016segmentation,yu2016modeling} involves taking a natural language description of a target region and returning the corresponding segmentation mask. Because the natural language expression can reference any concept, the task is open-vocabulary and the target can be a single instance, a coherent group of instances, or an entire semantic category, such as in \emph{all roads in the image patch} or \emph{the vegetation strip along the river}. The remote-sensing literature coined the term Referring Remote Sensing Instance Segmentation (RRSIS)~\cite{yuan2023rrsis} for a particular type of referring expression segmentation tasks in aerial imagery, with early datasets focusing on single instances. Later datasets like NWPU-Refer~\cite{yang2024large} expanded to group-level coverage. Referring expression segmentation is especially demanding in aerial images because top-down perspectives compress object scales, the use of color is not consistent, spatial resolution varies across sensors, many targets of interest occupy only a handful of pixels, objects frequently appear partially occluded, and the scenes themselves contain extreme object densities.

\begingroup
\setlength{\intextsep}{6pt}
\setlength{\abovecaptionskip}{2pt}
\setlength{\belowcaptionskip}{0pt}
\begin{figure}[t!]
\centering
\includegraphics[width=\columnwidth]{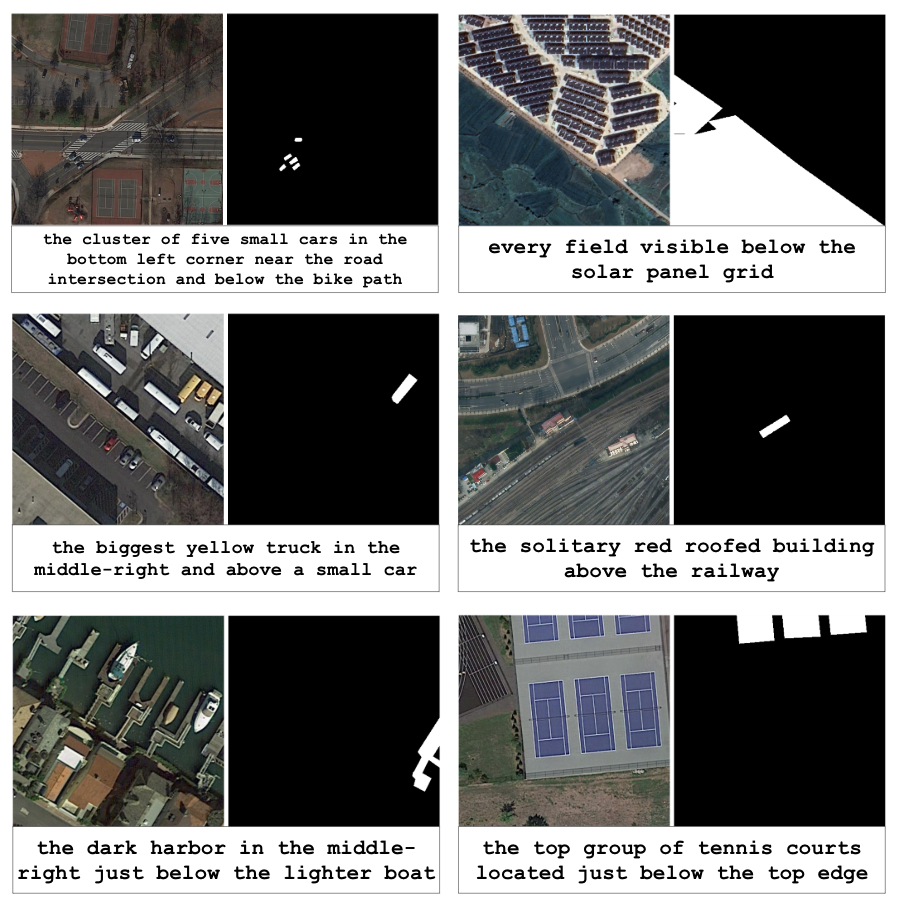}
\caption{Representative examples from the Aerial-D dataset, showing diverse types of referring expressions paired together with the corresponding aerial images and the ground truth segmentation masks.}
\label{fig:dataset_examples}
\end{figure}
\endgroup

A critical component for developing effective models for referring expression segmentation on aerial photographs is the access to high-quality datasets containing aerial imagery of different types, precise segmentation masks, and natural referring expressions using a rich vocabulary and covering different types of targets. To address this need, this work presents Aerial-D as a new large-scale referring expression segmentation dataset for aerial imagery, comprising 1,522,523 expressions across 37,288 images, which is significantly larger than prior RRSIS benchmarks~\cite{yuan2023rrsis,liu2024rotated,yang2024large}. Figure~\ref{fig:dataset_examples} highlights how this corpus spans rural and urban scenes, different types of objects and land coverage types, and both instances or groups of multiple objects, while retaining unrestricted and richly worded referring expressions tailored to each target.

To investigate how this data can translate into a practical referring expression segmentation model, we adopted RSRefSeg~\cite{chen2025rsrefseg} as a baseline architecture for our experiments. We train RSRefSeg on Aerial-D together with other existing datasets, namely RefSegRS~\cite{yuan2023rrsis}, RRSIS-D~\cite{liu2024rotated}, NWPU-Refer~\cite{yang2024large}, and Urban1960SatSeg~\cite{hao2025urban1960satseg}, aiming at a single model that can respond to language queries across both contemporary and historic imagery, achieving generalized referring expression segmentation on aerial photos.

The key contributions of this work include: (1) a comprehensive set of tools that enables the production of referring expression datasets from existing instance/semantic segmentation datasets, including a rule-based generator, methods for refining the expressions based on Large Language Models (LLMs), and image data augmentations; (2) the Aerial-D dataset comprising over 1.5 million expressions across 37,288 aerial images, created entirely through the proposed set of tools; and (3) a unified model trained on Aerial-D alongside four additional datasets, delivering high-quality referring expression segmentation over instances, groups, land coverage and general semantic classes, while maintaining reliable performance on both modern high-resolution images and also on degraded historical imagery typical of archival aerial surveys.

\section{Related Work}
\label{sec:related}

This section reviews datasets and recent architectural developments associated with aerial image understanding.

\subsection{Aerial Imagery Segmentation Datasets}

Reliable progress in aerial image understanding depends on datasets, for model training and evaluation, that capture individual objects (e.g., ships, vehicles) and semantic classes (e.g., roads, water, vegetation), as well as datasets that test language-based selection of specific targets.

LoveDA~\cite{wang2021loveda} is one of the richest datasets for semantic segmentation of aerial images, featuring both urban and rural environments and a total of six land-cover classes. Covering 536.15 km$^2$ with 0.3m-resolution imagery, LoveDA enables domain adaptation research by highlighting differences between built environments dominated by artificial structures and rural regions characterized by natural elements.

In turn, the iSAID dataset~\cite{zamir2019isaid} established the foundation for instance segmentation in aerial imagery, providing 655,451 object instance annotations across 15 categories in 2,806 high-resolution images. Building upon the DOTA dataset~\cite{xia2018dota}, iSAID addressed the unique challenges of aerial imagery, including high object densities, pronounced scale variations, and arbitrary orientations.

Research on remote sensing referring expression segmentation began with RefSegRS~\cite{yuan2023rrsis}, which introduced 4,420 image--language--mask triplets and formalized the Referring Remote Sensing Instance Segmentation (RRSIS) problem. The dataset highlighted specific challenges associated to aerial images, namely the small and densely packed targets or the cluttered layouts, where competent natural language processing is crucial for disambiguating between visually similar instances.

RRSIS-D~\cite{liu2024rotated} expanded both scale and annotation efficiency with 17,402 image--caption--mask triplets, generated through a semi-automated pipeline using the Segment Anything Model~\cite{sam} (SAM) to generate segmentation masks. The imagery in this dataset originated from the DIOR dataset~\cite{li2020dior}, grounding RRSIS-D in a broad remote-sensing corpus while leveraging automated mask generation. Beyond size, RRSIS-D targets several aerial-specific phenomena, namely broad spatial scales and diverse object orientations across 20 categories and seven attribute dimensions, enabling richer evaluation of language-guided selection in overhead scenes.

NWPU-Refer~\cite{yang2024large} further broadened coverage with 15,003 high-resolution images and 49,745 annotated targets spanning more than 30 countries. In contrast to semi-automated pipelines, NWPU-Refer emphasized purely manual annotation quality and explicitly supports single-object, multi-object, and non-object scenarios across 32 categories.

All the aforementioned datasets feature modern RGB images, although collected through different procedures and at different resolutions. Despite the many interesting potential applications, few previous studies have focused on the analysis of historical aerial imagery, and no previous studies have specifically focused on referring expression segmentation. Analyzing historical aerial photographs introduces practical complications, e.g. related to reduced contrast, grayscale capture, film artifacts, and geometric distortions. Urban1960SatSeg~\cite{hao2025urban1960satseg} is one of the few existing exceptions, featuring professionally annotated semantic segmentation masks for four classes over 1,240 km$^2$ of declassified mid-20th-century imagery. By concentrating on degraded visual conditions, this dataset provides a reference point for methods that must remain robust when handling archival aerial data.

Beyond Urban1960SatSeg, Sertel et al.~\cite{sertel2023historic} assembled a benchmark of historic panchromatic aerial photographs spanning populous regions in Turkey and Bulgaria from the 1950s through the 1970s. The author also demonstrated the suitability of the resulting dataset for land-use/land-cover segmentation by training U-Net++ and Deeplabv3 models with carefully tuned backbones. The results show that contemporary deep architectures can recover reliable land coverage maps from monochrome archival imagery, underscoring the viability of deep learning on historic aerial photographs.

\subsection{Neural Network Architectures for RRSIS}

Neural architectures for referring expression segmentation in aerial imagery combine vision backbones, language encoders, and fusion mechanisms that translate textual cues into pixel-level masks. Several proposals have been advanced in the recent literature. For instance, the Rotated Multi-Scale Interaction Network (RMSIN)~\cite{liu2024rotated} is a representative design that builds on a Swin Transformer~\cite{swin} visual encoder and a BERT~\cite{bert} language backbone. It also features an intra-scale interaction module that refines fine-grained details with Transformer blocks, a cross-scale interaction module that aligns multi-resolution features through cross-attention, and an adaptive rotated convolution module that injects rotation-aware convolutional filters to handle arbitrary object orientations.

MRSNet~\cite{yang2024large} also adopts a Swin--BERT backbone pairing, but alters how features interact. Instead of RMSIN's triplet of modules, this architecture instead employs an hierarchical fusion approach that first consolidates information within each scale before exchanging context across scales, leading to a more progressive flow of visual detail toward the mask decoder. The shift in interaction pattern highlights how architectural variants largely differ in their feature fusion strategies rather than wholesale backbone changes.

\begin{figure*}[t]
\centering
\includegraphics[width=0.9\textwidth]{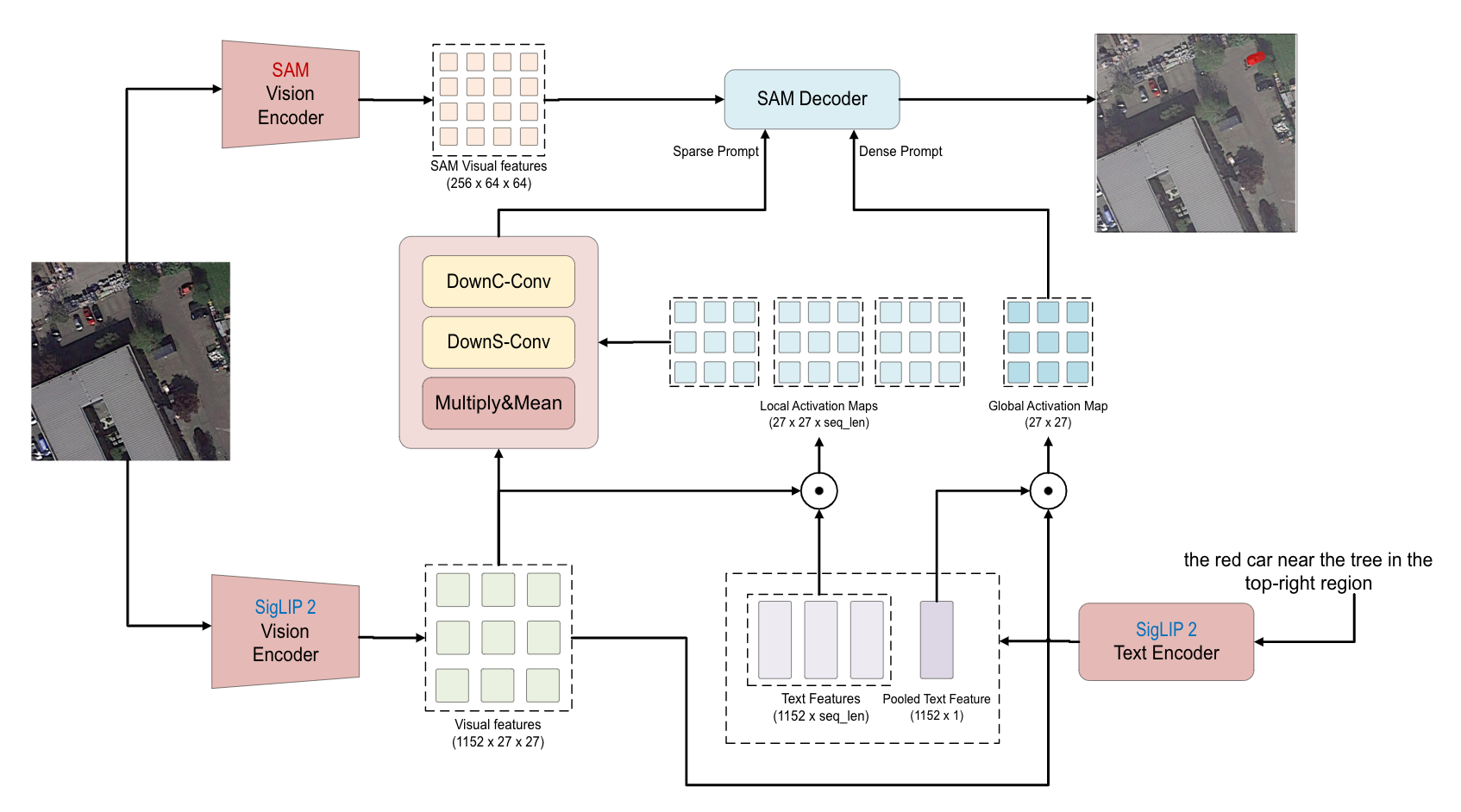}
\caption{Overview of the RSRefSeg architecture~\cite{chen2025rsrefseg}, which couples a pre-trained vision--language encoder with a segmentation decoder via a learned bridge.}
\label{fig:rsrefseg_arch}
\end{figure*}

RSRefSeg~\cite{chen2025rsrefseg} illustrates the alternative of leaning fully on large vision--language models. It replaces separately trained encoders with CLIP~\cite{clip} or SigLIP~\cite{siglip} dual encoders for multimodal feature extraction, at the same time also integrating SAM as the segmentation decoder. A lightweight AttnPrompter module converts language-conditioned features into sparse and dense prompts for SAM, while LoRA adapters fine-tune both the CLIP/SigLIP encoders and the SAM vision encoder to aerial imagery. Figure~\ref{fig:rsrefseg_arch} illustrates this configuration and highlights how the prompts interface with SAM. The original work presents two model scales: RSRefSeg-B pairs AttnPrompter with SAM-ViT-Base, whereas RSRefSeg-L instead uses SAM-ViT-Large. In our work, we retain this layout, reimplementing the architecture in PyTorch and training both variants on Aerial-D together with other external datasets, using originally published hyperparameters. The only change is that we double the LoRA rank for the SigLIP and SAM encoders in RSRefSeg-L, to handle the larger multi-dataset training schedule.

\section{Aerial-D Dataset Construction}
\label{sec:approach}

This section details our approach to constructing Aerial-D, i.e. a large-scale referring expression segmentation dataset for aerial imagery. Our methodology combines rule-based expression generation with a LLM-based expression refinement component, aiming at both scale and linguistic diversity. 

\subsection{Sources of Data}

The Aerial-D dataset was constructed from two complementary aerial image segmentation datasets with different annotation styles, namely iSAID with instance segmentation annotations, and LoveDA with semantic segmentation annotations. Together, these source datasets feature 21 distinct classes, with iSAID contributing 15 instance categories (e.g., vehicles, ships, or buildings) and LoveDA contributing 6 semantic land coverage classes (e.g., farmland, forest, or water).

Independently of the source, images in Aerial-D have a uniform size of $480\times480$ pixels, which keeps small iSAID objects large enough to describe and segment, while fitting the input expectations of the common vision encoders used in our model (i.e., the CLIP/SigLIP image encoding towers and SAM segmentation backbones \cite{clip,siglip,sam}).

We first resize the $1024\times1024$ LoveDA images directly to $480\times480$ pixels while preserving the semantic masks. iSAID has uniquely high-resolution imagery with varying aspect ratios, so we instead slide a $480\times480$ window, with a stride of 384 pixels across each source image, and keep the tiles that contain valid segmentation instances. After these resizing steps, we run connected-component analysis on LoveDA to turn buildings and water into pseudo-instance targets, noting that these categories tend to appear as isolated structures or bounded water bodies, thus making them natural candidates for creating additional instance-level descriptions. The remaining land coverage classes (e.g., farmland or forest) behave as contiguous surfaces, so we keep them as semantic regions and describe them holistically as in the example \emph{all agricultural land in the image}.

\subsection{Rule-Based Expression Generation}

After consolidating the images and annotations from both source datasets, we introduce the rule-based stage that produces deterministic referring expressions for the annotation masks. This stage first requires a precise definition of the targets that are to be described: (1) individual object instances; (2) compact same-category clusters obtained by running DBSCAN~\cite{ester1996density} on individual object regions, measuring their minimum edge-to-edge distance, and capping each cluster at eight members; and (3) class-level groups that aggregate every instance sharing a label within the patch.

\begin{figure*}[t]
\centering
\begin{minipage}{0.42\textwidth}
\centering
\includegraphics[width=0.7\textwidth]{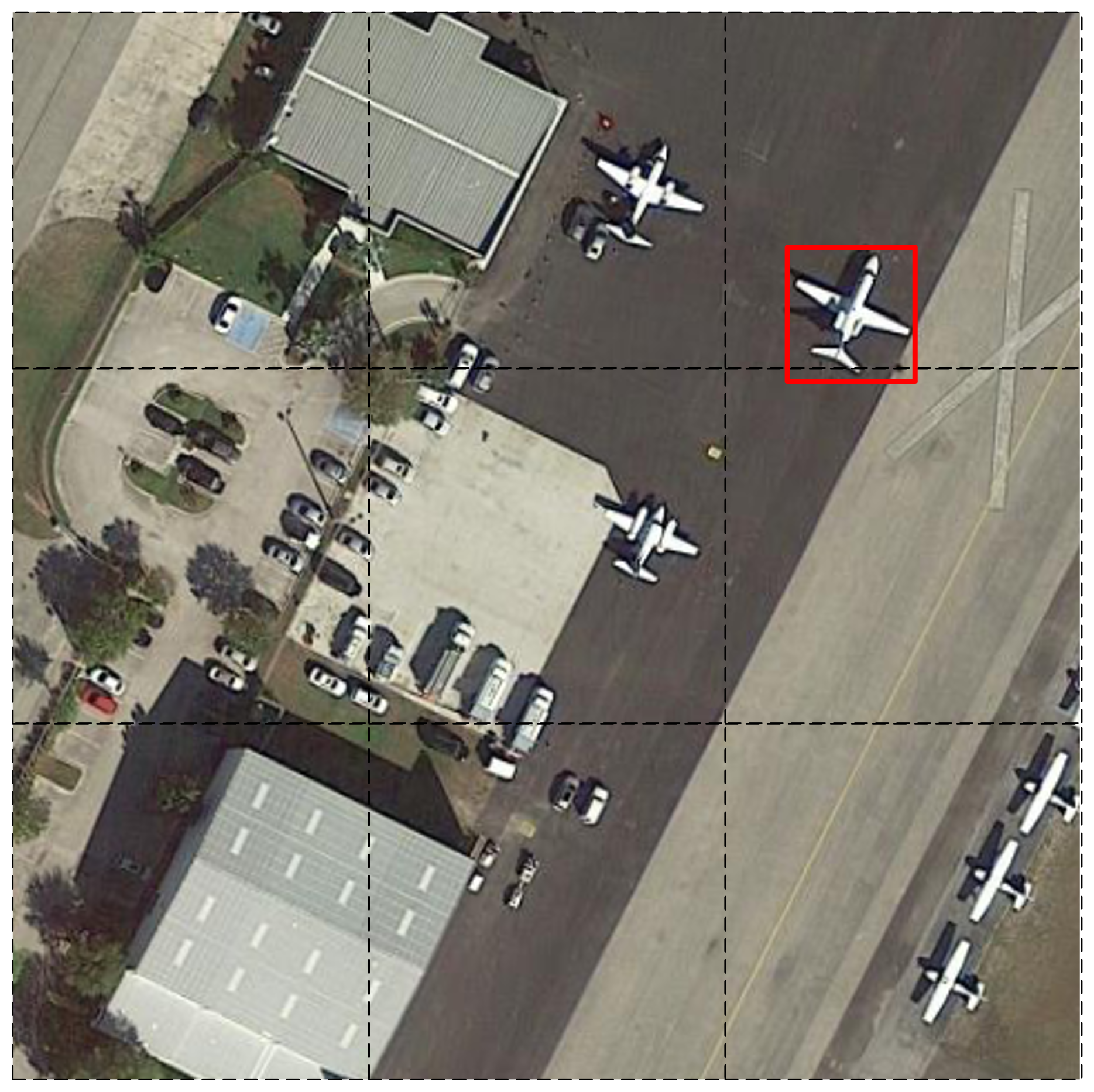}
\end{minipage}%
\begin{minipage}{0.58\textwidth}
\centering
\hspace{-0.5cm}
\raisebox{-0.2\height}{%
\footnotesize
\begin{tabular}{@{}p{3cm}p{7cm}@{}}
\toprule
\textbf{Rule Type} & \textbf{Example Instance} \\
\midrule
Category & \emph{plane} \\
Grid Position & \emph{in the top-right} \\
Extreme Position & No \\
Color Classification & \emph{light} \\
Directional Relations & \emph{to the bottom-right of a plane} \\
& \emph{to the top-right of a plane} \\
\midrule
\multicolumn{2}{l}{\textbf{Final Expressions}} \\
\multicolumn{2}{l}{\emph{the plane in the top-right}} \\
\multicolumn{2}{l}{\emph{the light plane in the top-right}} \\
\multicolumn{2}{l}{\emph{the plane in the top-right to the bottom-right of a plane}} \\
\multicolumn{2}{l}{\emph{the light plane in the top-right to the bottom-right of a plane}} \\
\multicolumn{2}{l}{\emph{the plane in the top-right to the top-right of a plane}} \\
\multicolumn{2}{l}{\emph{the light plane in the top-right to the top-right of a plane}} \\
\bottomrule
\end{tabular}%
}

\end{minipage}
\caption{An example illustrating rule-based generation for a single instance. The highlighted plane in the top-right section demonstrates how the system assigns spatial, visual, and relational cues that will later be combined into referring expressions.}
\label{fig:rule_example}
\end{figure*}

The core challenge in building Aerial-D relates to figuring out how to describe the target objects using only what we know from the segmentation masks and categories. We use a set of rules to extract different types of cues, from which an expression is then generated. We first use bounding box coordinates derived from the masks to understand where each object sits within the complete image patch. As shown in Figure \ref{fig:rule_example}, we divide each patch into a three-by-three grid marked with dotted lines, so we can say that an object is \emph{in the top-right} or \emph{in the center}. When we have multiple objects of the same type, we also check if any of them are in extreme positions, like the topmost or leftmost instance of that category.

From the masks for each object instance, we can also analyze the corresponding pixel colors by looking at Hue--Saturation--Value (HSV) distributions, to distinguish between light and dark objects, and between a small controlled palette of chromatic colors. We require at least 70\% dominance for achromatic labels ("light" or "dark") and a single hue to occupy at least 60\% of the chromatic pixels, before we commit to a specific color for an object. When neither threshold is met, no color descriptor is assigned and the cue is discarded. These thresholds allow us to handle ambiguous multi-hue regions, and help us ignore noisy signals that would otherwise mislead the language generation. We also avoid using color descriptors for buildings and water, since these typically show mixed colors that we consider to not be useful for identification.

We also create relationships between nearby objects by calculating angles between their positions, allowing us to generate expressions like \emph{the ship to the left of the harbor} or \emph{the vehicle above the building}. We consider eight directional relationships, namely above, below, to the left of, to the right of, and the four diagonal directions.

All the aforementioned rules combine to generate various referring expressions for each object instance, as demonstrated in Figure \ref{fig:rule_example} where a single plane generates multiple possible descriptions, including \emph{the plane in the top-right}, \emph{the light plane in the top-right}, and versions with relational descriptions. However, a significant challenge emerges when multiple objects end up with identical characteristics from which we then generate the exact same expressions, creating ambiguous references where one phrase could describe multiple different objects. We solve this problem by taking the set of all expressions for every target in each image, matching them to find duplicates, and discarding any referring expression that appears more than once as ambiguous.

\subsection{LLM Expression Generation}
\label{subsec:llm_expression_generation}
\begin{figure*}[t]
\centering
\begin{minipage}{0.42\textwidth}
\centering
\includegraphics[width=0.7\textwidth]{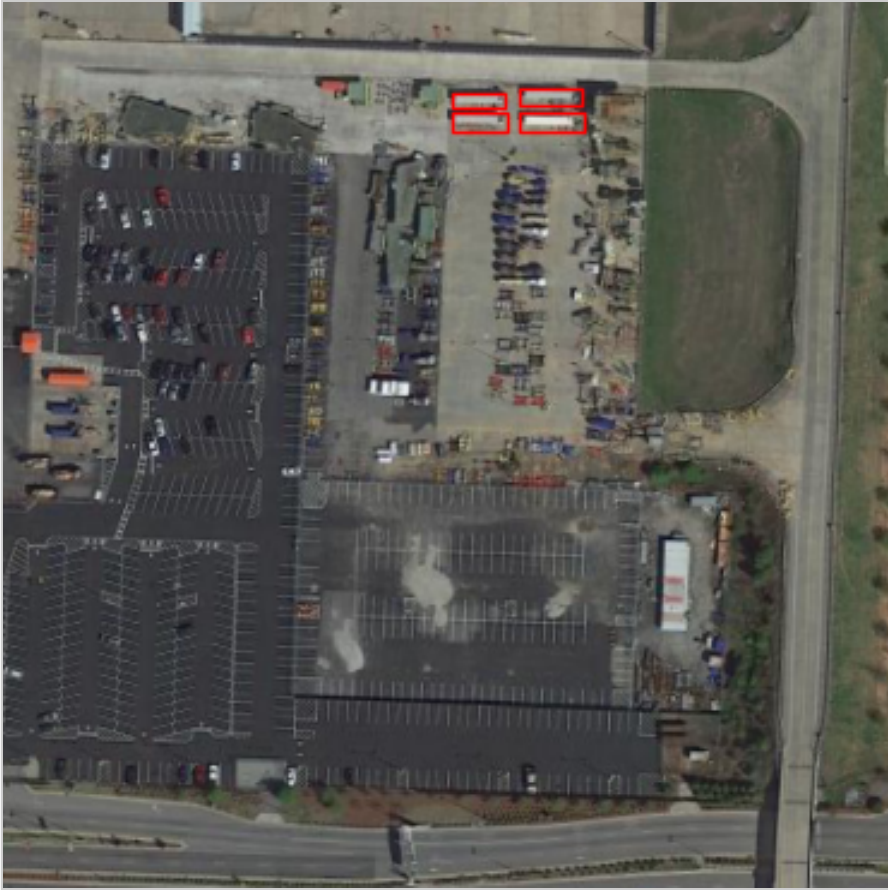}
\end{minipage}%
\begin{minipage}{0.58\textwidth}
\centering
\hspace{-0.5cm}
\raisebox{-0.2\height}{%
\footnotesize
\begin{tabular*}{\linewidth}{@{}p{0.30\linewidth}@{\extracolsep{\fill}}p{0.68\linewidth}@{}}
\toprule
\textbf{Expression Type} & \textbf{Example} \\
\midrule
Rule-Based & \emph{the group of 4 large vehicles in the top-center} \\
\midrule
LLM Language Variation & \emph{the cluster of four big vehicles at the upper-middle region} \\
\midrule
LLM Visual Variation & \emph{the four large vehicles lined up side by side just below the pale paved strip at the very top-middle} \\
\midrule
LLM Visual Variation & \emph{the set of four big vehicles parked in a single row in the upper-center beside the grassy area to the right} \\
\bottomrule
\end{tabular*}%
}
\end{minipage}
\caption{An example illustrating the complete expression generation and enhancement pipeline. Starting from the aerial image (left), the rule-based approach generates initial expressions, which the LLM then refines into language variations and visual variations that incorporate additional contextual details (right).}
\label{fig:llm_enhancement_example}
\end{figure*}

While rule-based expression generation provides a solid foundation, the expressions that are produced have significant limitations in language variation and visual detail coverage, lacking the ability to reference contextual elements beyond predefined source dataset categories. To address these limitations, we employ a multimodal Large Language Model (LLM) to enhance our dataset, providing both images and expressions as inputs to the model, and asking it to rewrite and improve the original referring expressions. We specifically prompt the LLM with two complementary tasks, as shown in Figure \ref{fig:llm_enhancement_example}. The first task focuses on linguistic variation, creating natural language alternatives for each rule-based expression, without heavy reliance on visual cues. The second task uses visual information, asking the model to examine surrounding features in the image around the target object.

We overlap the target region with red bounding boxes to guide the model during the first task, and pair each prompt with a focused close-up crop so that small or dense targets stay visible. For land coverage categories that lack crisp bounding boxes, we supply a dual-image prompt consisting of a masked overlay and the clean image, which helps the model anchor the relevant region. This combination of bounding box overlays, dual images, and mask prompts keeps the enhancement grounded on the correct area of the scene.

The dual-task prompting strategy transforms basic expressions, like \emph{the group of 4 large vehicles in the top center}, into linguistically diverse alternatives, such as \emph{the cluster of four big vehicles at the upper middle region} and visually detailed descriptions like \emph{the four large vehicles lined up side by side just below the pale paved strip at the top-center}, as shown in Figure~\ref{fig:llm_enhancement_example}. The model identifies and references contextual elements not captured in the original datasets, such as the \emph{pale paved strip} and the \emph{grassy area}.

While we noted that a frontier LLM such as OpenAI o3 could produce highly accurate results, the application to a very large dataset produced with our rule-based approach would be prohibitive. For instance, our full dataset from the rule-based stage contains 259,709 captured targets, including both objects and groups. To generate refined expressions we require processing each target individually, leading to at least 259,709 separate LLM requests, and an estimated cost of thousands of dollars with OpenAI o3, as outlined in Table \ref{tab:cost_comparison}. We therefore considered a knowledge distillation~\cite{hinton2015distilling} approach, which uses 500 high-quality outputs from OpenAI o3 on a representative random subset of targets from the initial dataset, as training data for supervised fine-tuning using the parameter-efficient QLoRA method~\cite{qlora} on Gemma3-12B~\cite{gemma_3_2025}.

During fine-tuning, we apply LoRA adapters across both the text decoder and the SigLIP vision encoder embedded in Gemma3, which improves instruction adherence, suppresses hallucinations, and stabilizes the two-task output schema. The custom-tailored Gemma3 model was then used to process all 259,709 targets on a single GPU, while honoring the dual-task prompt structure. Notably, a qualitative inspection of the produced results showed that the distilled model's output quality approaches that of OpenAI o3, once fine-tuned. The qualitative comparisons in Figure \ref{fig:distillation_comparison} show closely matched enhancements, with markedly reduced hallucinations relative to the base Gemma3 model.

\subsection{Historic Image Filter Augmentation}
\label{subsec:historic_filters}

\begin{figure*}[t]
\centering
\begin{minipage}[b]{0.24\textwidth}
\centering
\includegraphics[width=0.75\textwidth]{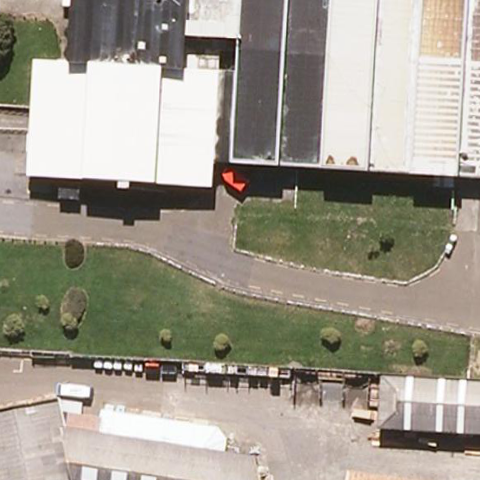}
\end{minipage}
\begin{minipage}[b]{0.24\textwidth}
\centering
\includegraphics[width=0.75\textwidth]{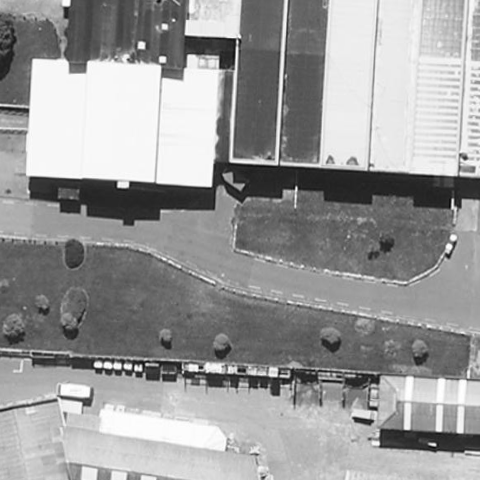}
\end{minipage}
\begin{minipage}[b]{0.24\textwidth}
\centering
\includegraphics[width=0.75\textwidth]{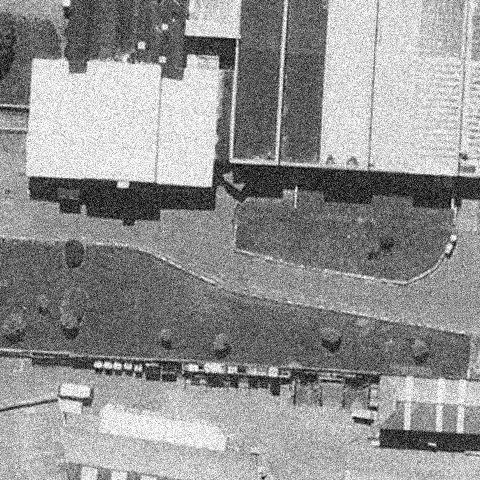}
\end{minipage}
\begin{minipage}[b]{0.24\textwidth}
\centering
\includegraphics[width=0.75\textwidth]{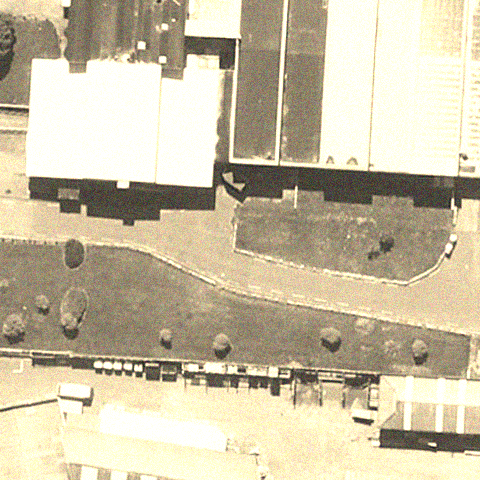}
\end{minipage}
\caption{Ilustration for the application of filters used for simulating historical images. The image shows the original RGB capture (far left), grayscale conversion, grayscale with grain, and sepia toning with sensor noise. Each variant preserves structure while introducing degradations representative of archival imagery.}
\label{fig:historic_filters}
\end{figure*}

In order to improve robustness when processing historical imagery, we propose training with three parametric transformations that reproduce characteristic degradations of historical aerial photographs (Figure~\ref{fig:historic_filters}). These filters are applied on the fly during training, rather than baked into the dataset, so that each mini-batch can include either a clean or an historically degraded view of the same image.

Let $I_{o}(x)\in[0,255]^3$ denote the RGB image contents at pixel $x$, and let $\operatorname{clip}(\cdot)$ clamp values to $[0,255]$. We simulate grayscale image captures by converting the RGB values to luminance, as shown in \eqnref{eq:gray}:
\begin{equation}
I_{\text{bw}}(x) = 0.299\,I_{o}(x)[R] + 0.587\,I_{o}(x)[G] + 0.114\,I_{o}(x)[B].
\label{eq:gray}
\end{equation}

To emulate film response and grain, we first apply a mild gamma adjustment as shown in \eqnref{eq:gamma}, then a linear contrast change around the mean as shown in \eqnref{eq:contrast}, and finally add Gaussian noise according to \eqnref{eq:grain}:
\begin{equation}
I_{\gamma}(x) = 255\,\big(I_{\text{bw}}(x)/255\big)^{\gamma}.
\label{eq:gamma}
\end{equation}
\begin{equation}
I_{c}(x) = \big(I_{\gamma}(x) - \mu\big)\,c + \mu.
\label{eq:contrast}
\end{equation}
\begin{equation}
I_{\text{g}}(x) = \operatorname{clip}\big(I_{c}(x) + \eta(x)\big),\quad \eta(x)\sim\mathcal{N}(0,\sigma^2).
\label{eq:grain}
\end{equation}
We use $\gamma=1.2$, $c=0.8$, and $\sigma=0.1\times 255$ to produce mild contrast loss and film grain.

Finally, we apply a fixed sepia transform defined according to \eqnref{eq:sepia}, followed by uniform sensor noise \eqnref{eq:sepia_noise}:
\begin{equation}
\begin{bmatrix} S(x) \end{bmatrix}
= \operatorname{clip}\left(
\begin{bmatrix}
0.272 & 0.534 & 0.131 \\
0.349 & 0.686 & 0.168 \\
0.393 & 0.769 & 0.189
\end{bmatrix}
\begin{bmatrix} I_{o}(x)[R] \\ I_{o}(x)[G] \\ I_{o}(x)[B] \end{bmatrix}
\right).
\label{eq:sepia}
\end{equation}
\begin{equation}
I_{\text{s}}(x) = \operatorname{clip}\big(\mathbf{S}(x) + \xi(x)\big),\quad \xi(x)\sim\mathcal{U}(0,50).
\label{eq:sepia_noise}
\end{equation}

These effects mimic tonal range reduction, film grain, and scanning artifacts typical of aerial photography, while preserving the spatial structure that segmentation relies on. Figure~\ref{fig:historic_filters} illustrates the visual impact of each transformation.

\subsection{Final Dataset Statistics}

\begin{figure}[!t]
\centering
\includegraphics[width=\columnwidth]{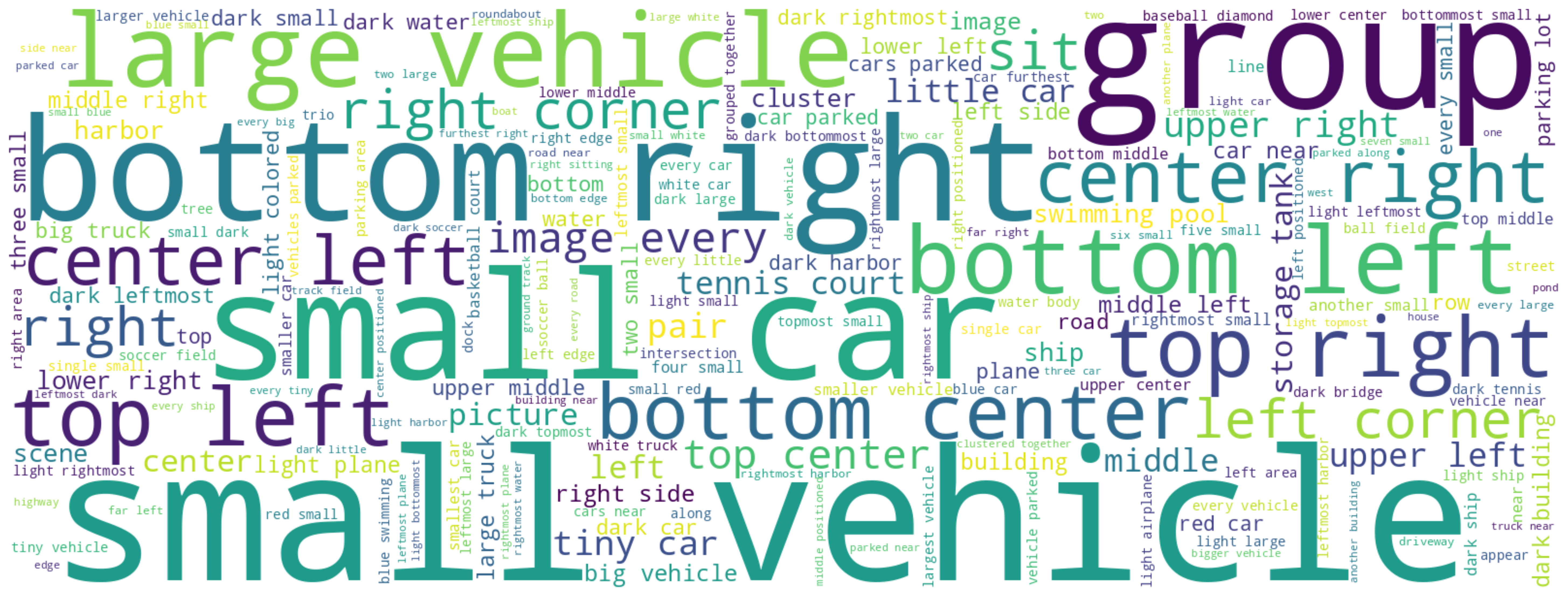}
\caption{Word cloud visualization for the most frequent terms in the Aerial-D referring expressions, highlighting the domain-specific vocabulary and spatial positioning descriptors that are characteristic of aerial imagery features.}
\label{fig:expression_wordcloud}
\end{figure}

The rule-based generation process yields 506,194 referring expressions and identifies 259,709 annotated targets across the considered data (Table~\ref{tab:llm_enhancement_stats}). Building on this base, the LLM enhancement procedure is prompted to produce one language variation for each original expression, and two unique visual-detail expressions for each target, respectively adding 496,895 and 519,434 expressions, and resulting in a total of 1,522,523 expressions. Of these, 1,278,453 expressions describe discrete instances or groups, while 244,070 reflect the land coverage classes that remain at the semantic level. Figure~\ref{fig:expression_wordcloud} illustrates the vocabulary used in the resulting set of expressions.

\begin{figure*}[t]
\centering
\begin{minipage}{0.48\textwidth}
\centering
\includegraphics[width=\textwidth]{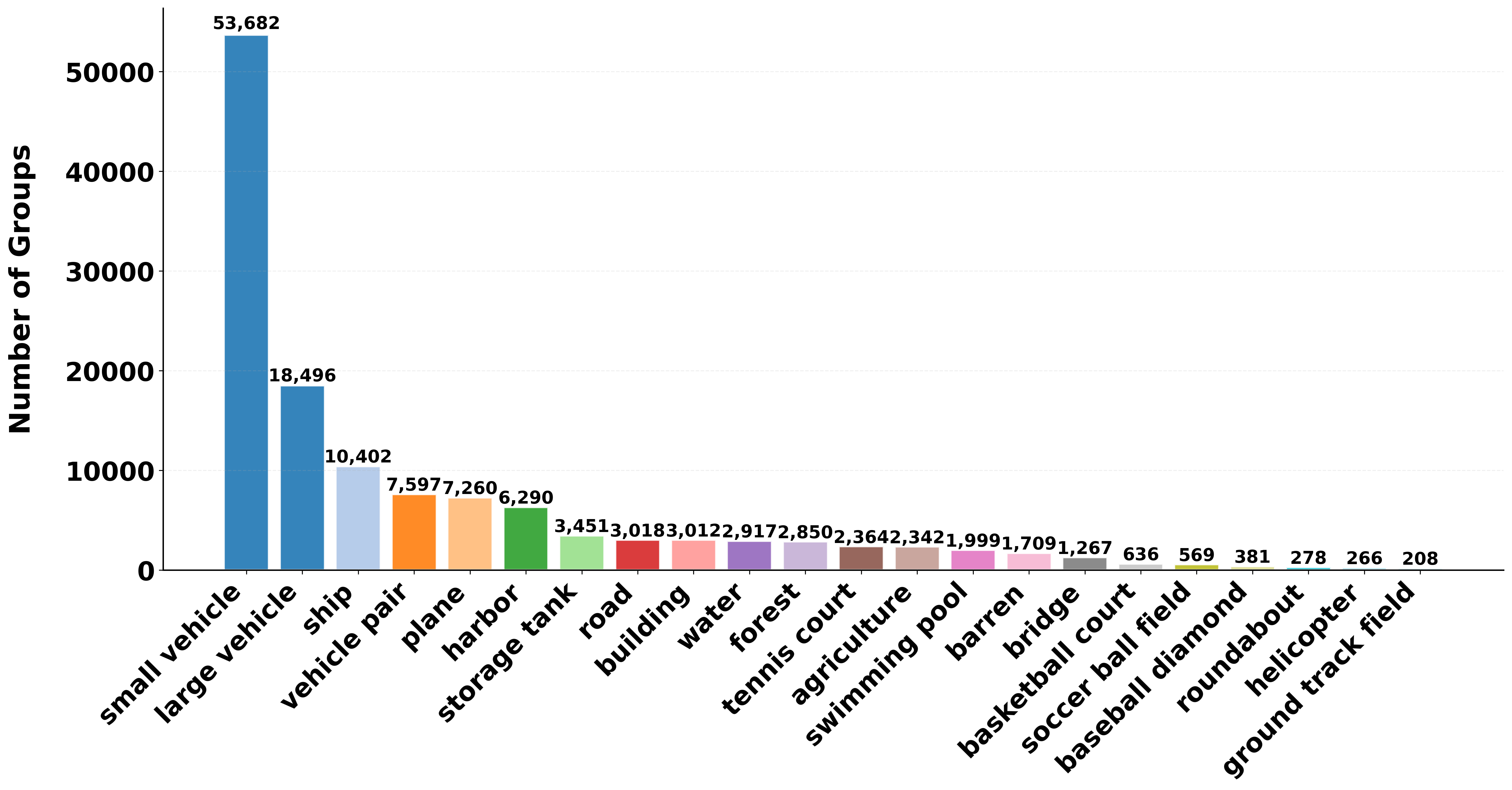}
\end{minipage}\hfill
\begin{minipage}{0.48\textwidth}
\centering
\includegraphics[width=\textwidth]{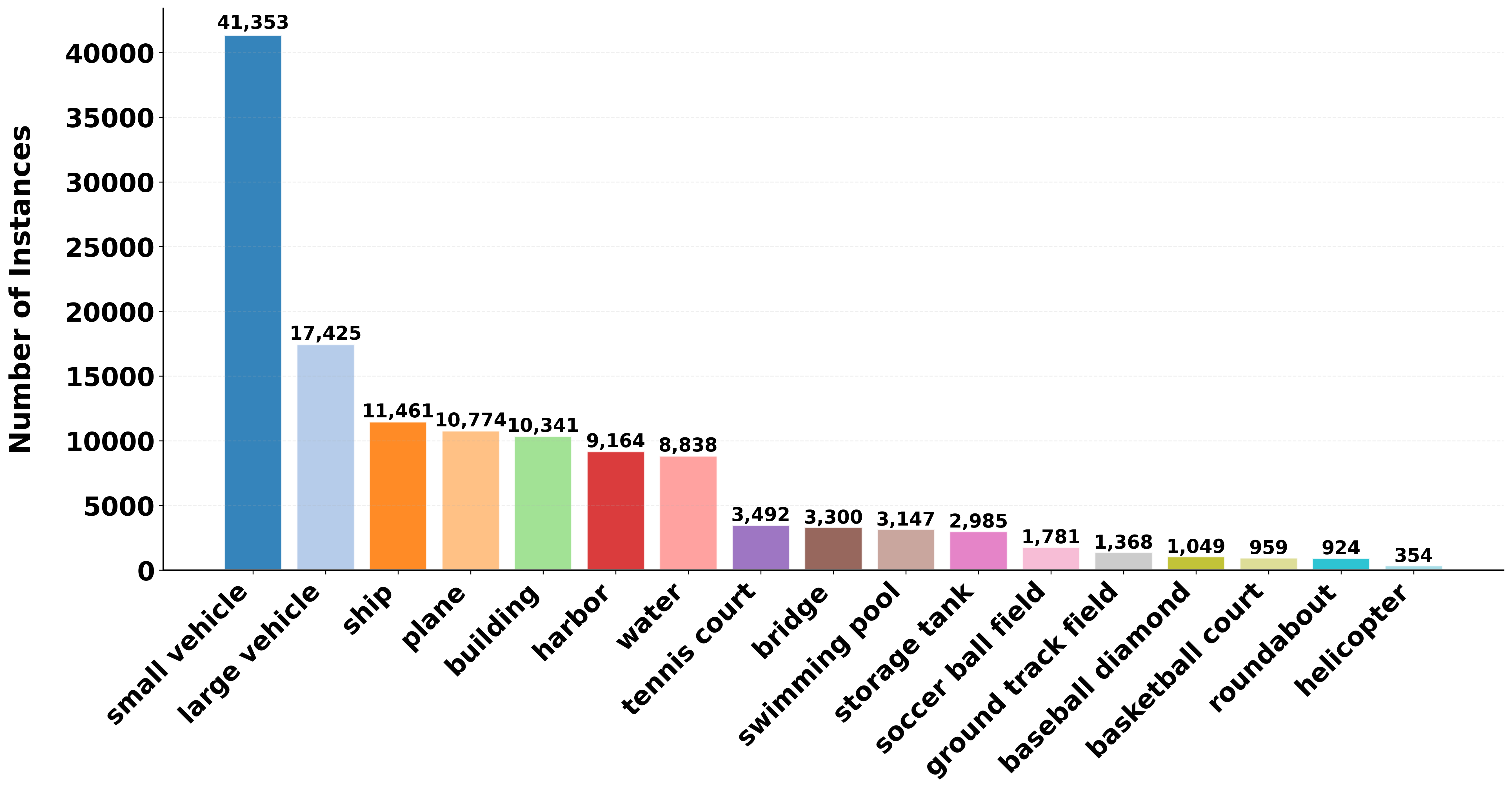}
\end{minipage}
\caption{An analysis on the distribution across categories for the referring expressions on the Aerial-D dataset. The left plot shows the distribution of grouped and semantic targets, showing the prevalence of region-level categories in the dataset. The right plot shows the overall distribution of individual instance annotations across semantic categories, demonstrating the dataset's coverage of aerial object types.}
\label{fig:category_distributions}
\end{figure*}

Table~\ref{tab:dataset_comparison} compares Aerial-D with prior RRSIS datasets, showing that Aerial-D contains nearly three times as many images as previous datasets. Each image typically includes many segmented targets, and each target is paired with multiple referring expressions. The table also clarifies how the corpus splits between the 1.28 million instance-level expressions and the 244 thousand semantic expressions, with these latter being absent from earlier RRSIS datasets. Beyond its scale, we note that Aerial-D relies on a fully automatic pipeline that can also be applied to more data sources, combining rule-based generation with LLM enhancement, supporting both single-object and multi-object references, and preserving the original category balance, as illustrated in Figure~\ref{fig:category_distributions}.

\begin{table*}[t]
\centering
\caption{Comparison between Aerial-D and other Existing RRSIS Datasets.}
\label{tab:dataset_comparison}
\resizebox{\textwidth}{!}{%
\begin{tabular}{@{}lcccccccc@{}}
\toprule
\textbf{Dataset} & \textbf{Image Resolution} & \textbf{Images} & \textbf{Instance Expr.} & \textbf{Semantic Expr.} & \textbf{Single-object} & \textbf{Multi-object} & \textbf{Image Size} & \textbf{Annotation Generation} \\
\midrule
RefSegRS & 0.13m & 4,420 & 4,420 & -- & \checkmark & $\times$ & 512 & Manual \\
RRSIS-D & 0.5m--30m & 17,402 & 17,402 & -- & \checkmark & $\times$ & 800 & Semi-automatic\\
NWPU-Refer & 0.12m--0.5m & 15,003 & 49,745 & -- & \checkmark & \checkmark & 1,024--2,048 & Manual \\
\midrule
\textbf{Aerial-D} & \textbf{0.3m--4.5m} & \textbf{37,288} & \textbf{1,278,453} & \textbf{244,070} & \textbf{\checkmark} & \textbf{\checkmark} & \textbf{480} & \textbf{Automated + LLM} \\
\bottomrule
\end{tabular}%
}
\end{table*}

\begin{table}[t]
\centering
\caption{Referring expression distribution by source and target type.}
\label{tab:llm_enhancement_stats}
\footnotesize
\begin{tabular}{@{}lrrr@{}}
\toprule
\textbf{Expression Source} & \textbf{Train} & \textbf{Test} & \textbf{Total} \\
\midrule
Rule-Based Expressions & 371,360 & 134,834 & 506,194 \\
LLM Language Variations & 364,396 & 132,499 & 496,895 \\
LLM Visual Variations & 382,038 & 137,396 & 519,434 \\
\midrule
Instances & 944,224 & 334,229 & 1,278,453 \\
Semantic Classes & 173,570 & 70,500 & 244,070 \\
\midrule
\textbf{Total Expressions} & \textbf{1,117,794} & \textbf{404,729} & \textbf{1,522,523} \\
\bottomrule
\end{tabular}
\end{table}

\section{Experiments}
\label{sec:experiments}

This section presents a comprehensive experimental evaluation for the usefulness of Aerial-D, spanning model training, cross-dataset generalization, and targeted ablations. We begin by outlining the RSRefSeg model used in our study and its training configuration. Then, we report cross-dataset results on established aerial referring expression segmentation benchmarks. Beyond aggregate performance, we also include: (i) ablations on the expression enhancement strategies; (ii) ablations on historic filter training; and (iii) qualitative comparisons across language models for expression enhancement (OpenAI o3, the base Gemma3 model, and our distilled Gemma3 model), coupled with a cost analysis of these alternatives.

\subsection{Model Architecture}
\label{subsec:model_architecture}

Evaluating the use of Aerial-D demands a segmentation model that preserves the link between natural-language instructions and precise masks, while handling densely packed aerial targets. RSRefSeg~\cite{chen2025rsrefseg} meets these requirements and it already demonstrated state-of-the-art results on the RRSIS-D~\cite{liu2024rotated} dataset. We reimplemented this architecture in PyTorch~\cite{pytorch} and verified the RRSIS-D results before extending the experiments to Aerial-D, which confirmed that the neural network design transfers reliably to new datasets.

Our reimplementation of RSRefSeg mirrors the original component pairing, with SigLIP2~\cite{siglip2} supplying the image--text encoder and SAM~\cite{sam} providing the mask decoder. We fine-tune both modules with Low-Rank Adaptation~\cite{lora} layers placed on the query and value projections of each vision Transformer~\cite{vit} encoder block, and on the query, key, value, and output projections of the text encoder. Two checkpoints appear throughout the experiments. RSRefSeg-B keeps SAM-ViT-Base and LoRA rank $r=16$ for a lighter configuration, whereas RSRefSeg-L upgrades to SAM-ViT-Large with rank $r=32$ to maximize capacity while preserving the same training recipe. In both cases, the SigLIP2 model corresponds to the \texttt{SigLIP2-SO400M} variant, as made available through HuggingFace, with an input resolution of 384x384 pixels.

\subsection{Experimental Setup}
\label{subsec:experimental_setup}

Experiments focused on assessing the usefulness of Aerial-D together with four other datasets in the area. Model training adhered to the original RSRefSeg recipe, so that eventual differences arise from the data mixture rather than from custom optimization. Batches contain four samples with gradient accumulation of two steps, yielding an effective batch size of eight instances. All experiments were executed on a single NVIDIA RTX~A6000 GPU. We employ AdamW~\cite{adamw} with a learning rate of $1\times10^{-4}$, weight decay of 0.01, polynomial decay with power of 0.9, mixed precision~\cite{mixedprecision} training, and gradient clipping at 1.0 to mirror the original training dynamics. During training, we also apply one of the three historic filters, selected with equal probability, to 20\% of the training images in each non-historic dataset. Note that the Urban1960SatSeg dataset is inherently composed of historic imagery, so it receives no additional filtering and provides direct supervision for historical conditions.

In order to keep the cross-dataset mixture balanced, Aerial-D contributes only with the \emph{LLM Visual Variations} subset highlighted in Section~\ref{subsec:ablation_studies}. An ablation shows that this subset carries the strongest signal while still covering every target. We thus have that limiting Aerial-D to these expressions keeps the millions of available sentences from overwhelming the other four public datasets, each of which supplying only tens of thousands of annotations. The combined run therefore spans Aerial-D (LLM Visual Variations), RRSIS-D~\cite{liu2024rotated}, NWPU-Refer~\cite{yang2024large}, RefSegRS~\cite{yuan2023rrsis}, and Urban1960SatSeg~\cite{hao2025urban1960satseg}.

Model testing follows the original dataset splits, and Aerial-D relies on the full test split (405K expressions), which is further broken down into instance targets and semantic regions, as shown in Table~\ref{tab:aeriald_variants}.

To probe robustness we also evaluate fully filtered test sets for Aerial-D, RRSIS-D, NWPU-Refer, and RefSegRS, by converting every image with one of the three historic filters. Historic scores appear in italics alongside the original scores.

\subsection{Evaluation Results}
\label{subsec:evaluation_results}

Tables~\ref{tab:aeriald_variants} and \ref{tab:combined_training_results} report results for the combined model trained jointly on all datasets. Both tables focus on the mean IoU~\cite{everingham2010pascal} and overall IoU evaluation metrics for the original test splits, alongside results from the filtered counterparts that simulate historical conditions.

Table~\ref{tab:aeriald_variants} isolates Aerial-D and reports performance by target type. Instance targets correspond to explicit objects or groups, while the "All Targets" column aggregates both instance-level and semantic-level references across the full 405K test expressions. These values serve as baseline checkpoints for future work that evaluates new neural network architectures on Aerial-D.

Across the external datasets featured in Table~\ref{tab:combined_training_results}, our RSRefSeg-B and RSRefSeg-L model checkpoints remain competitive with previously published results, despite being trained within a single unified pipeline. This comparison is particularly noteworthy because the published models were trained on individual target datasets, which avoids interference from multi-dataset mixing and does not include historic filter augmentation. In contrast, our training recipe blends five datasets, applies historic filters to more than 20\% of the training images, and supervises both instance-level and semantic-level targets drawn from Aerial-D. The historic columns illustrate that the LoRA-tuned RSRefSeg-L model retains accuracy under simulated degradations, closing the gap between clean and historic imagery without requiring dataset-specific adjustments. Table~\ref{tab:combined_training_results} also foregrounds how the larger model checkpoint strengthens overall IoU on the more recent NWPU-Refer dataset, while maintaining the strong RRSIS-D performance reported in prior work.

The performance gap on the RefSegRS dataset deserves particular attention. While the previous RMSIN model achieves 59.96\% mIoU, RSRefSeg-L reaches 44.52\%. This gap reflects an important distribution shift, as RefSegRS contains referring patterns (e.g., vehicles along specific roads that are referenced through expressions in the plural form) that differ significantly from Aerial-D, RRSIS-D, and NWPU-Refer. Multi-dataset training biases the resulting models toward the majority distribution, causing lower performance on the out-of-distribution, and in some cases conflicting, RefSegRS samples.

Overall, the combined evaluation shows competitive results against published studies across every previous dataset, while providing reproducible Aerial-D baselines. The modest gap between the original and historic images confirms that injecting filtered imagery during training delivers robustness without eroding accuracy on contemporary photographs.

\begin{figure}[t]
\centering
\begin{minipage}[b]{0.48\columnwidth}
\centering
\includegraphics[width=\textwidth]{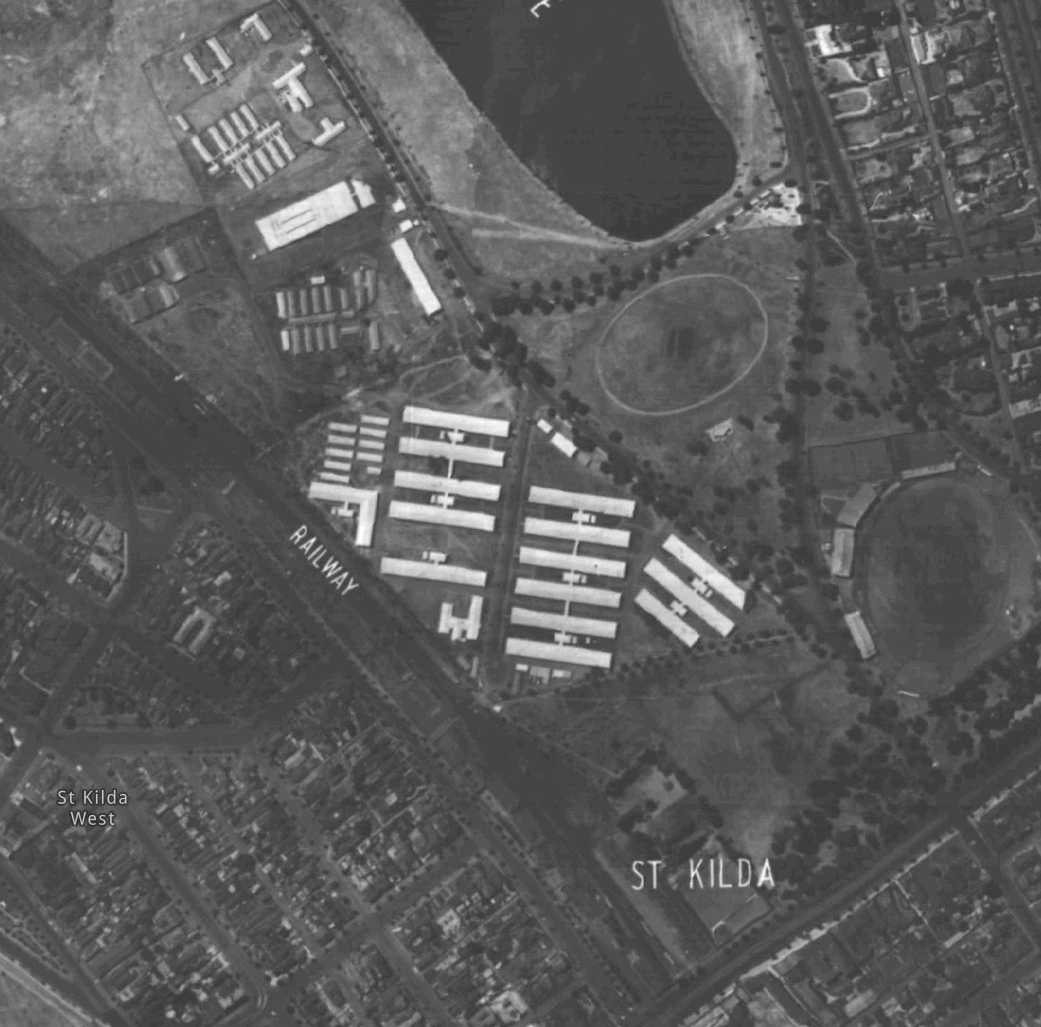}
\end{minipage}\hfill
\begin{minipage}[b]{0.48\columnwidth}
\centering
\includegraphics[width=\textwidth]{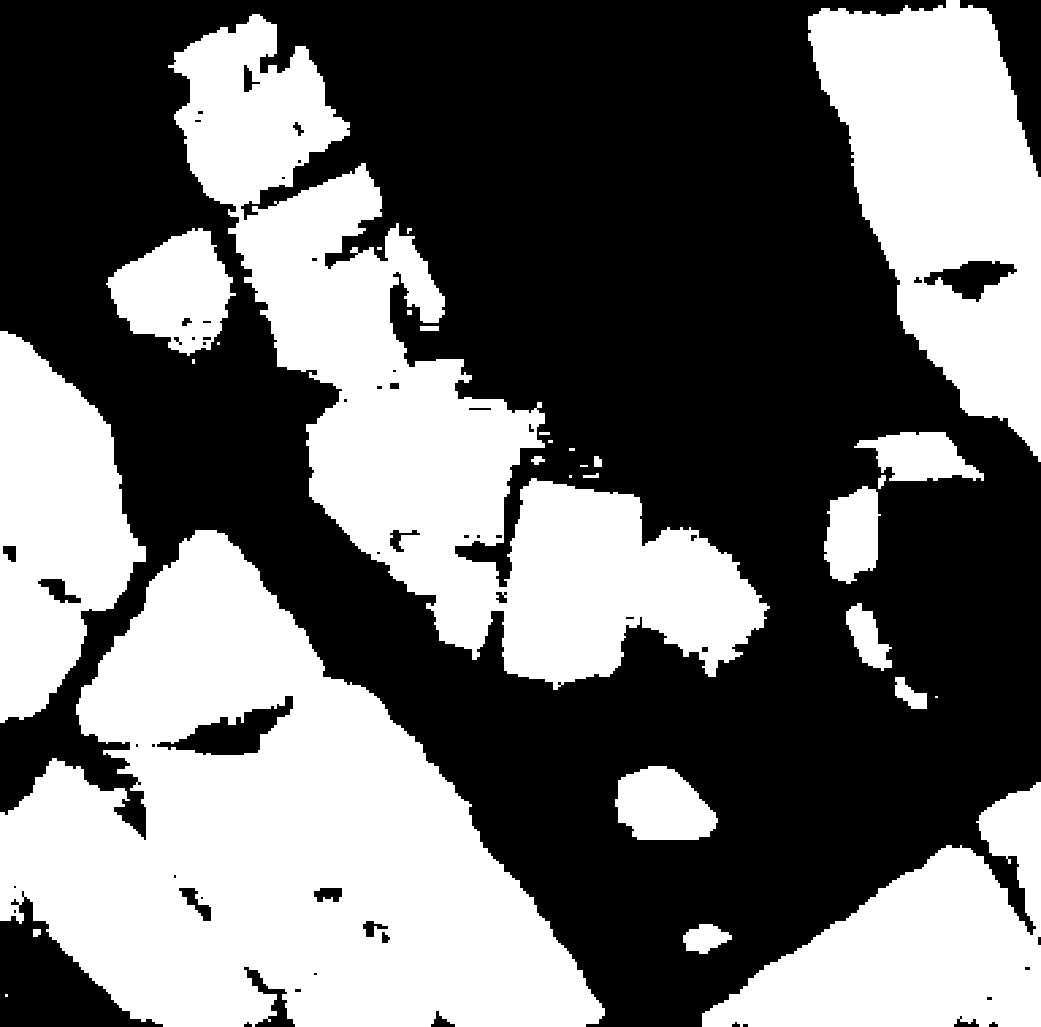}
\end{minipage}\\[4pt]
\begin{minipage}[b]{0.48\columnwidth}
\centering
\includegraphics[width=\textwidth]{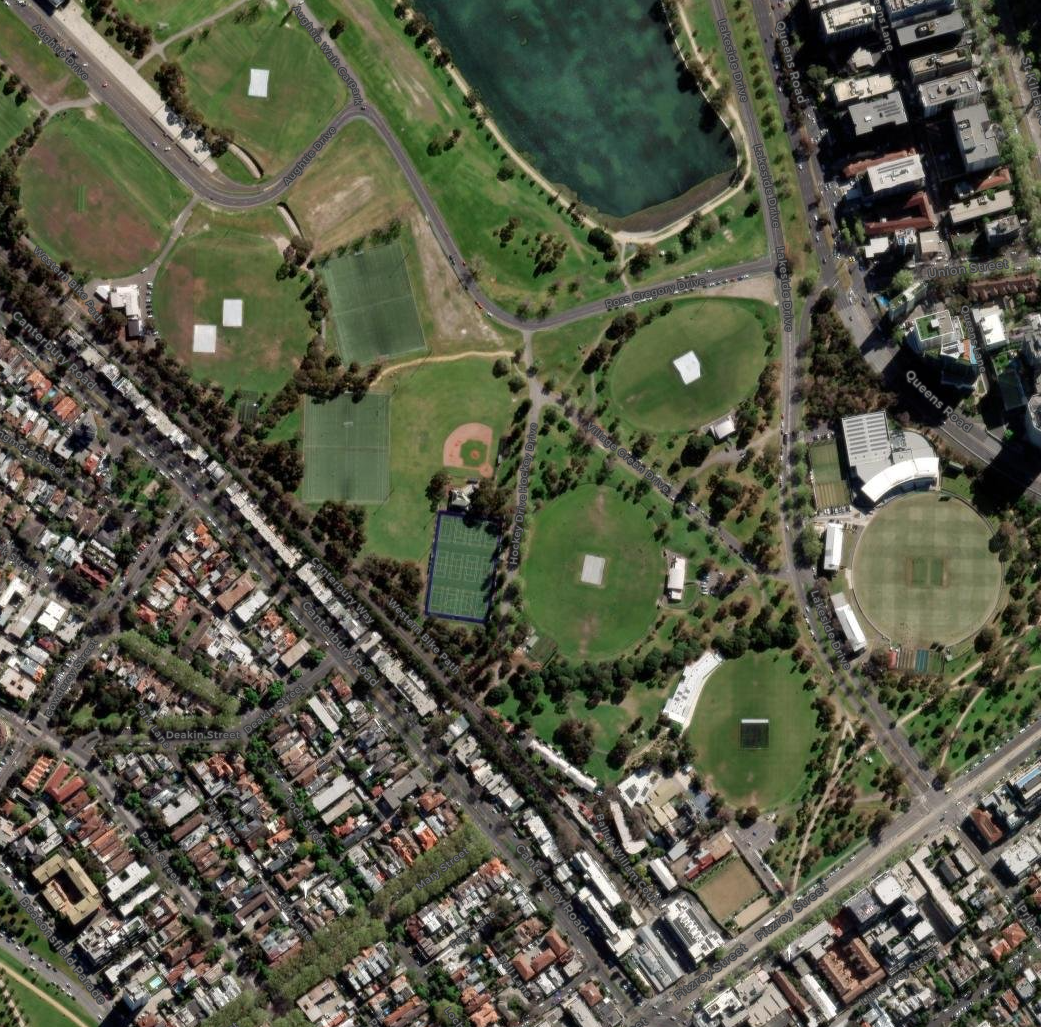}
\end{minipage}\hfill
\begin{minipage}[b]{0.48\columnwidth}
\centering
\includegraphics[width=\textwidth]{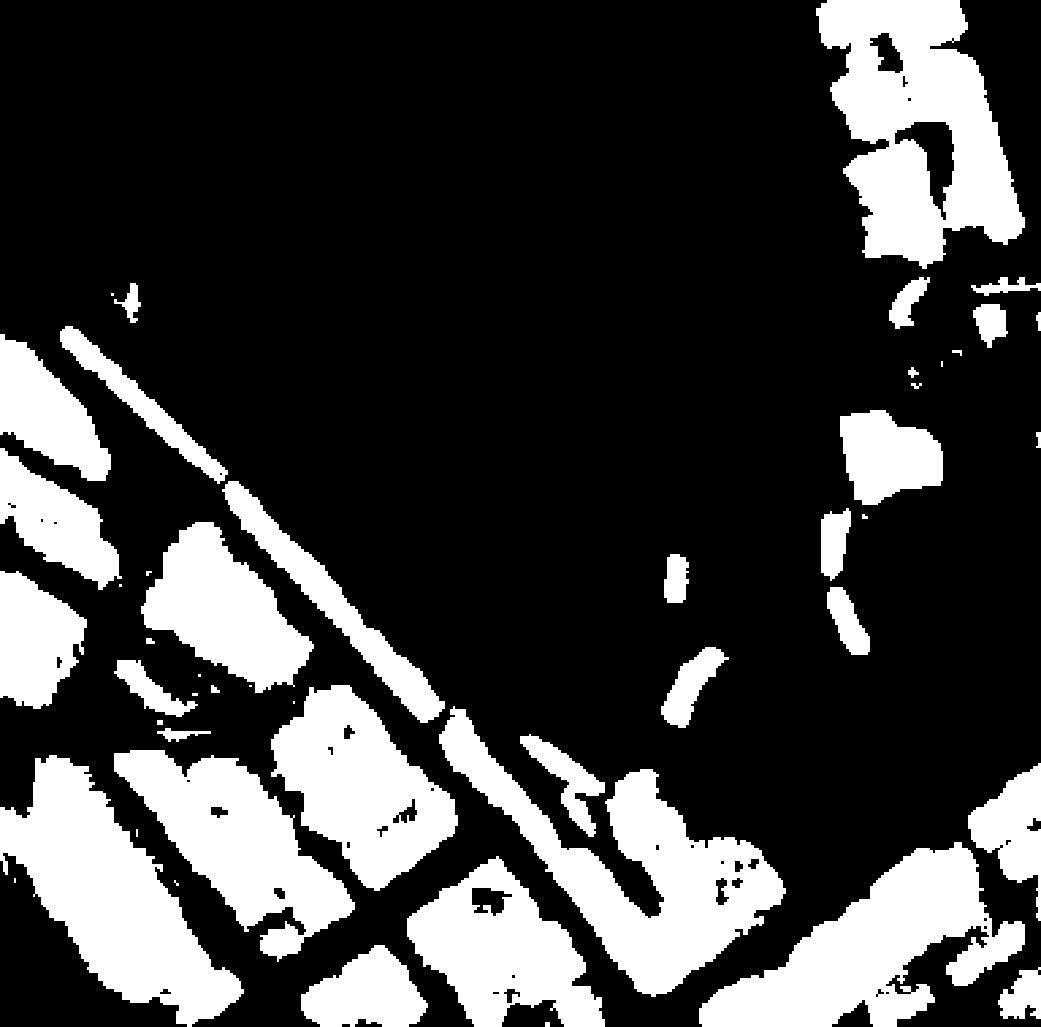}
\end{minipage}
\caption{An image depicting the Melbourne riverside district and showing RSRefSeg tracking \emph{all building areas in the image} across a 1945 grayscale photograph (top), and across a contemporary RGB capture (bottom). Left: original images; right: predicted segmentation masks.}
\label{fig:melbourne_appendix}
\end{figure}

\begin{figure}[t]
\centering
\begin{minipage}[b]{0.48\columnwidth}
\centering
\includegraphics[width=\textwidth]{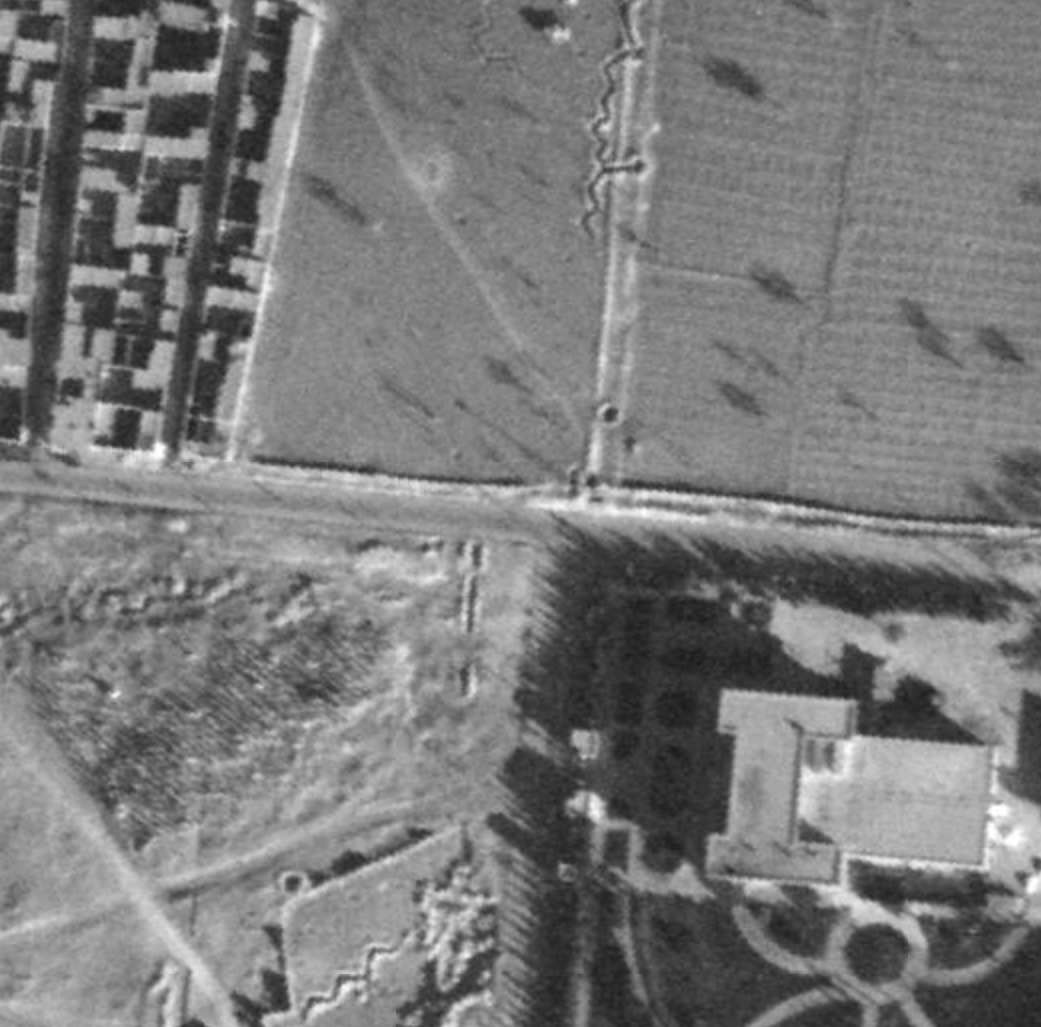}
\end{minipage}\hfill
\begin{minipage}[b]{0.48\columnwidth}
\centering
\includegraphics[width=\textwidth]{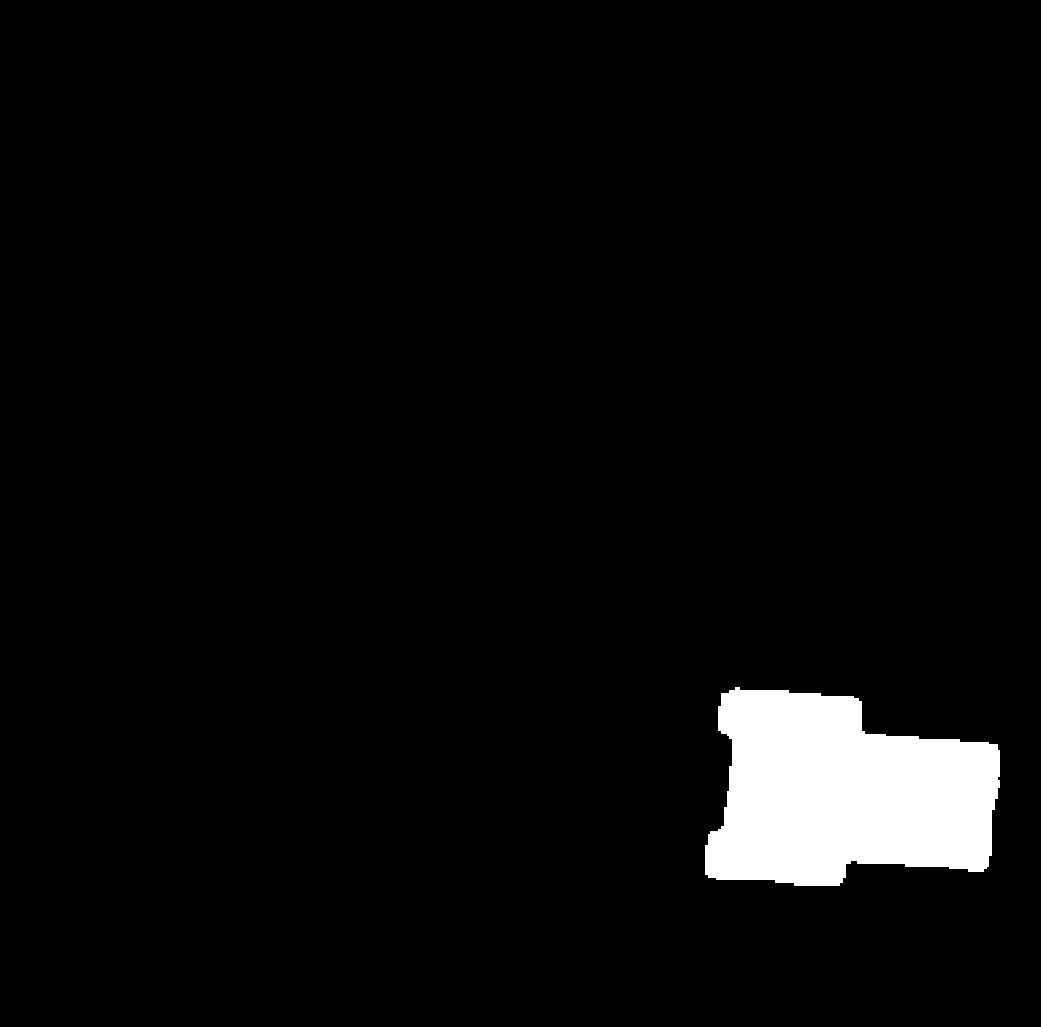}
\end{minipage}\\[4pt]
\begin{minipage}[b]{0.48\columnwidth}
\centering
\includegraphics[width=\textwidth]{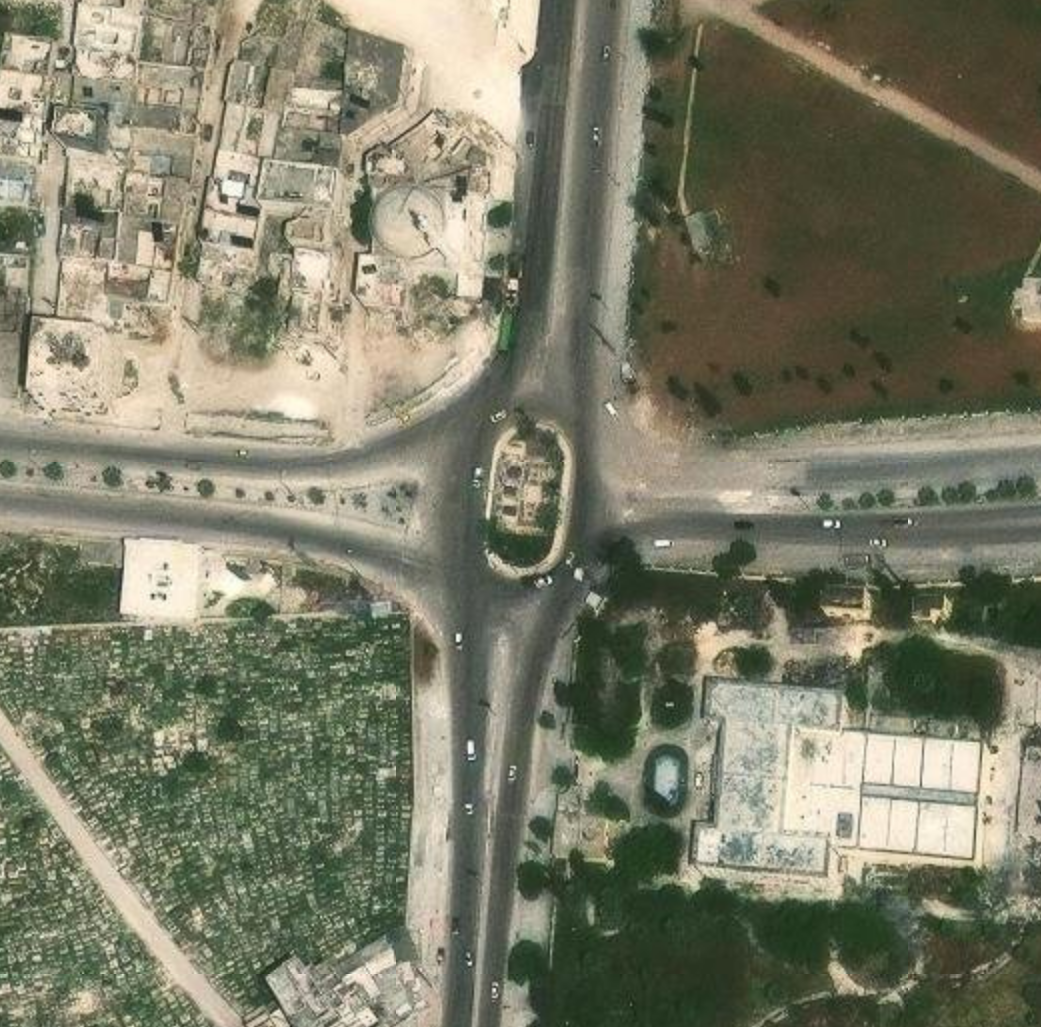}
\end{minipage}\hfill
\begin{minipage}[b]{0.48\columnwidth}
\centering
\includegraphics[width=\textwidth]{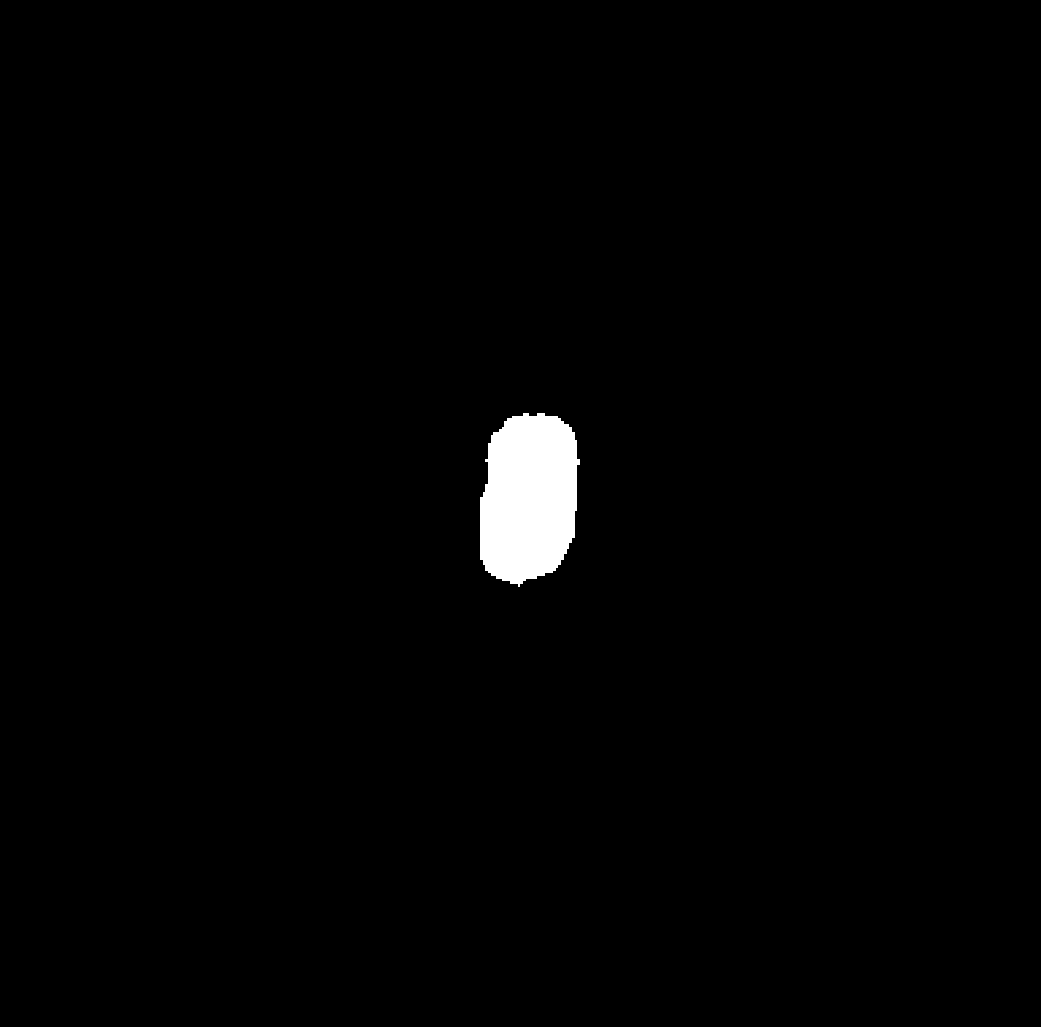}
\end{minipage}
\caption{An image depicting Aleppo, in Syria, and showing RSRefSeg adapting to two distinct expressions across U-2 reconnaissance footage (top row: \emph{the building at the bottom-right}) and a modern capture (bottom row: \emph{the roundabout in the center intersection}). Left: original images; right: predicted segmentation masks.}
\label{fig:u2_appendix}
\end{figure}

\begin{figure}[t]
\centering
\begin{minipage}[b]{0.48\columnwidth}
\centering
\includegraphics[width=\textwidth]{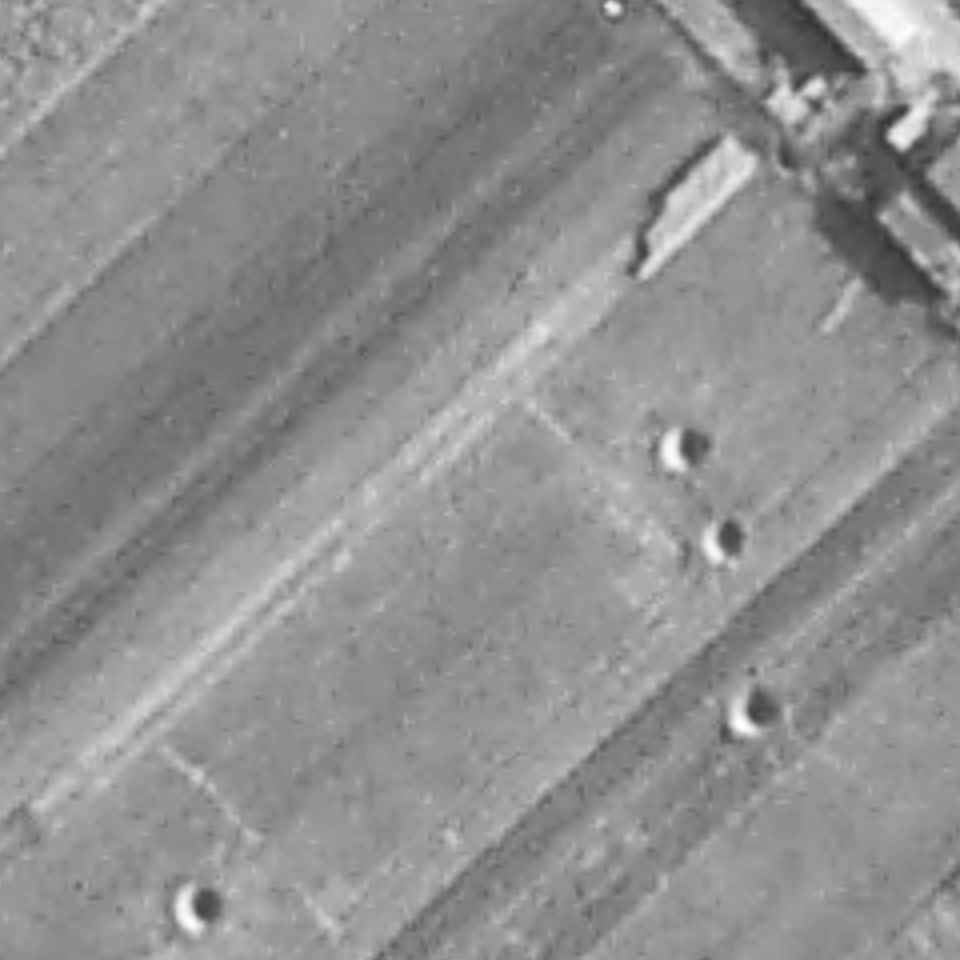}
\end{minipage}\hfill
\begin{minipage}[b]{0.48\columnwidth}
\centering
\includegraphics[width=\textwidth]{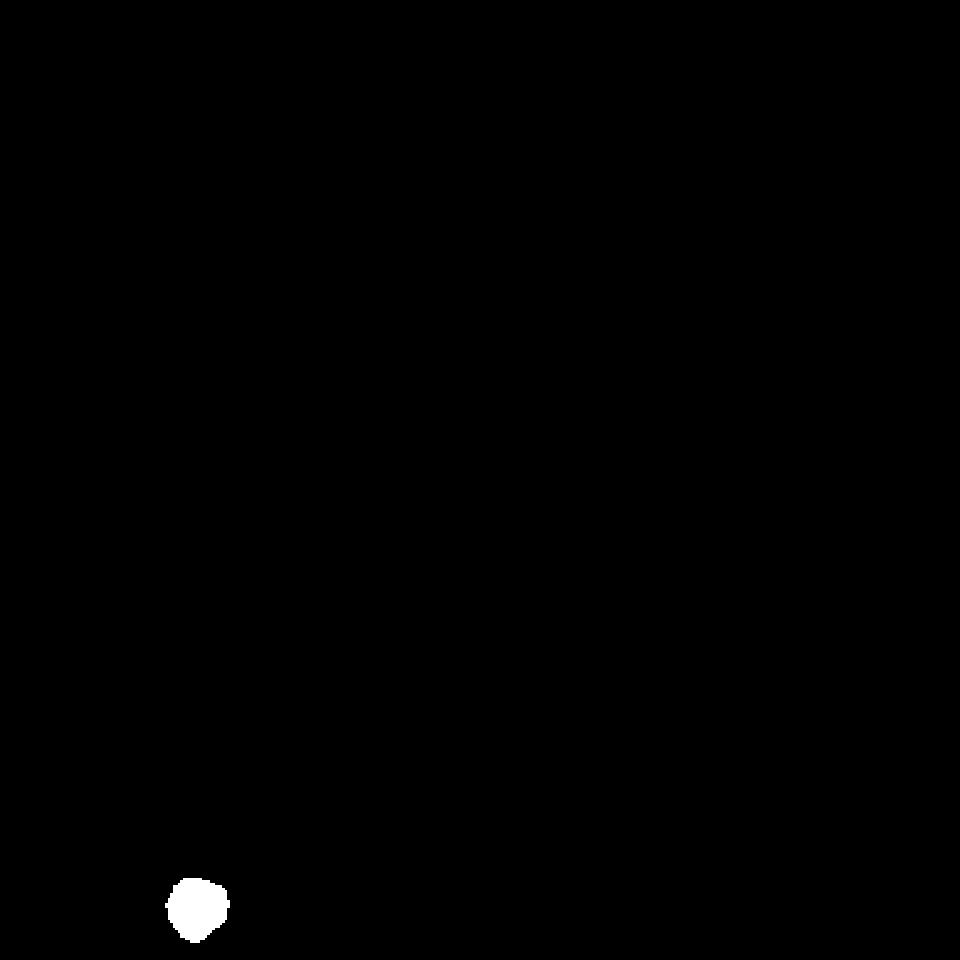}
\end{minipage}\\[4pt]
\begin{minipage}[b]{0.48\columnwidth}
\centering
\includegraphics[width=\textwidth]{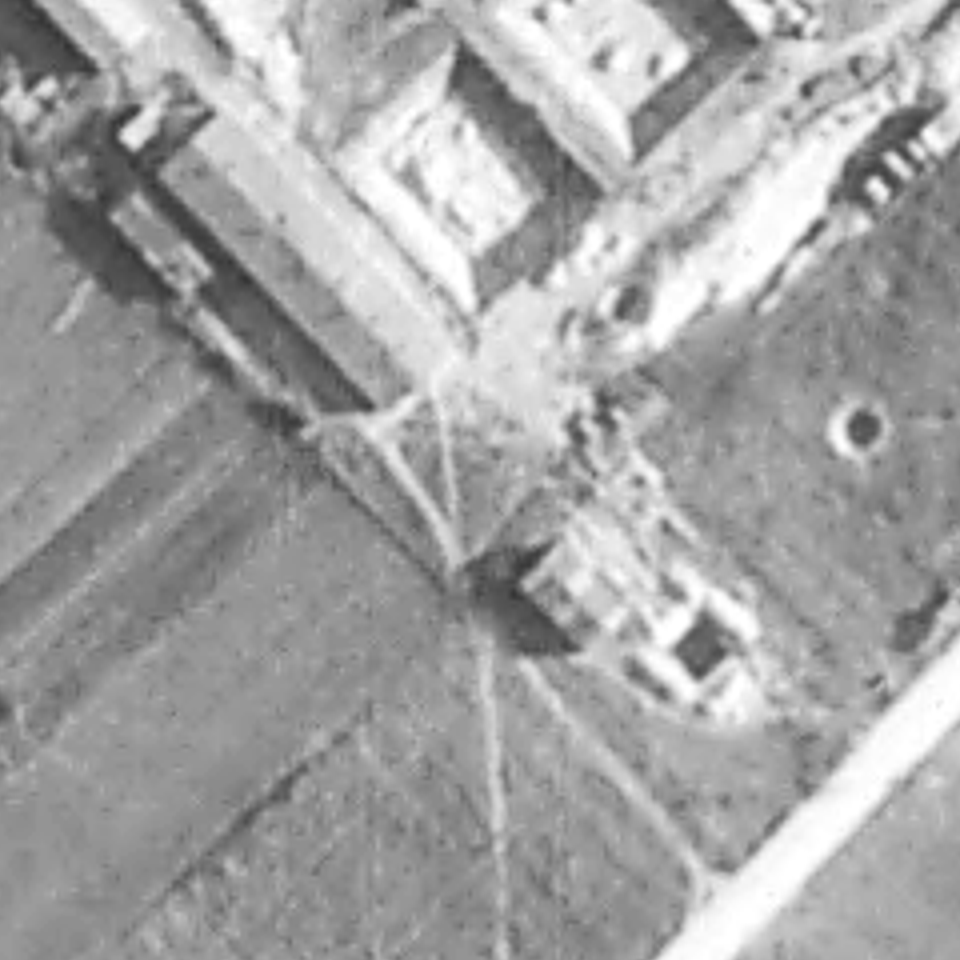}
\end{minipage}\hfill
\begin{minipage}[b]{0.48\columnwidth}
\centering
\includegraphics[width=\textwidth]{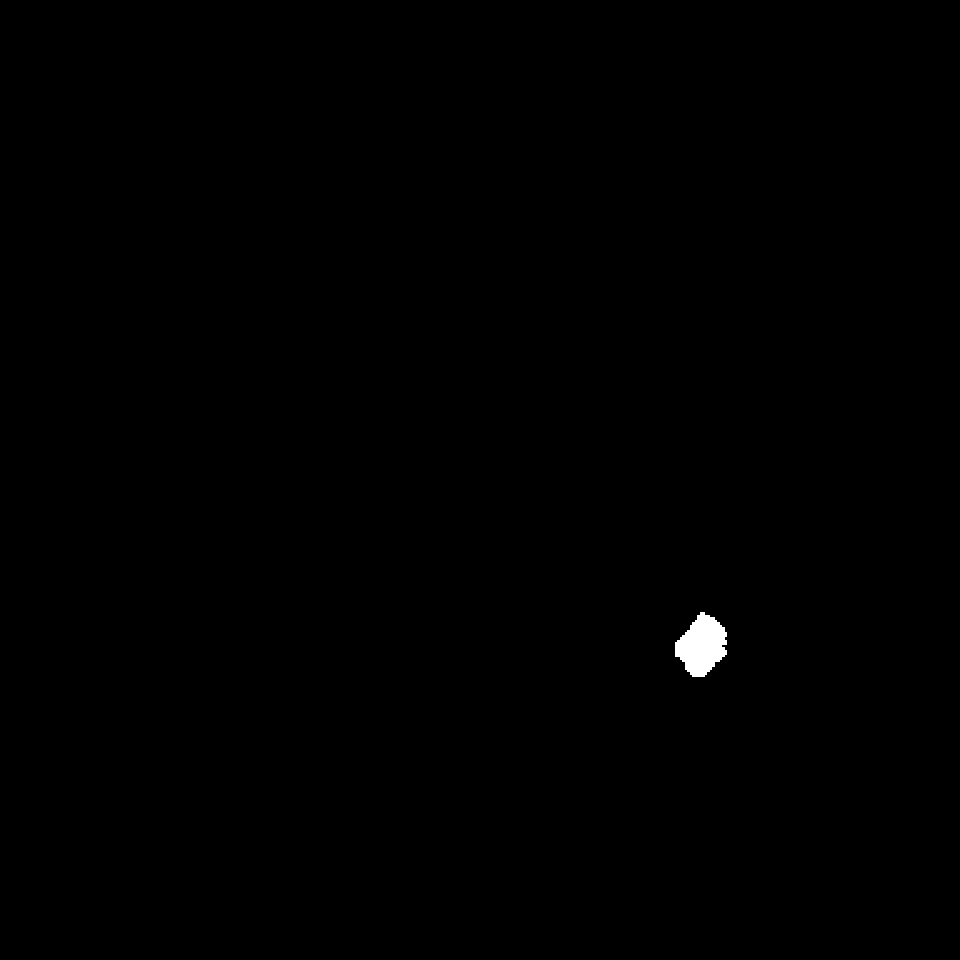}
\end{minipage}
\caption{Zero-shot generalization on images from the CHAI WWII archive. Top row: \emph{the crater to the bottom-left of a group of 3 craters}; bottom row: \emph{the square hole next to the road}. Despite never encountering craters during training, the model follows the relational spatial instructions learned from Aerial-D. Left: original 1944 imagery; right: predicted segmentation masks.}
\label{fig:chai_generalization}
\end{figure}

Beyond the quantitative results, Figures~\ref{fig:melbourne_appendix}, \ref{fig:u2_appendix}, and \ref{fig:chai_generalization} demonstrate qualitative performance on real historical imagery. Figure~\ref{fig:melbourne_appendix} shows the model segmenting \emph{all building areas in the image} across both a 1945 grayscale photograph and a contemporary RGB capture of Melbourne's riverside district~\cite{melbourne1945}, enabling tracking of urban change over decades. Figure~\ref{fig:u2_appendix} evaluates the model on Aleppo, Syria~\cite{HammerUr2019}, using two distinct expressions (i.e., \emph{the building in the bottom-right} and \emph{the roundabout in the center intersection}) on U-2 reconnaissance imagery and the corresponding modern counterpart. Most notably, Figure~\ref{fig:chai_generalization} demonstrates zero-shot generalization on an image from the CHAI WWII archive~\cite{marvin2023chai}, where the model correctly segments crater fortifications despite never encountering such targets during training. The model follows complex relational expressions such as \emph{the crater to the bottom-left of a group of three craters} and \emph{the square hole next to the road}, confirming that the spatial relationships learned from Aerial-D can transfer to entirely unseen object categories.

\begin{table*}[t]
\centering
\caption{Baseline performance of RSRefSeg variants on the Aerial-D validation split. Historic scores are shown in italics.}
\label{tab:aeriald_variants}
\resizebox{\textwidth}{!}{%
\begin{tabular}{@{}l|cc|cc|ccccc@{}}
\toprule
\multirow{2}{*}{\rule{0pt}{2.6ex}\textbf{Model}} & \multicolumn{2}{c|}{\textbf{Instance Targets}} & \multicolumn{2}{c|}{\textbf{Semantic Targets}} & \multicolumn{5}{c}{\textbf{All Targets}} \\
\cmidrule(lr){2-3} \cmidrule(lr){4-5} \cmidrule(lr){6-10}
 & \textbf{mIoU} & \textbf{oIoU} & \textbf{mIoU} & \textbf{oIoU} & \textbf{mIoU} & \textbf{oIoU} & \textbf{Pass@0.5} & \textbf{Pass@0.7} & \textbf{Pass@0.9} \\
\midrule
RSRefSeg-B & 45.63\% & 60.54\% & 54.07\% & 64.10\% & 47.10\% / \emph{39.77\%} & 62.40\% / \emph{56.60\%} & 53.78\% & 36.15\% & 6.60\% \\
RSRefSeg-L & 48.66\% & 62.97\% & 55.28\% & 64.82\% & \textbf{49.81\%} / \emph{42.18\%} & \textbf{63.29\%} / \emph{58.41\%} & \textbf{57.66\%} & \textbf{40.13\%} & \textbf{7.67\%} \\
\bottomrule
\end{tabular}%
}
\end{table*}

\begin{table*}[t]
\centering
\caption{Cross-dataset validation results for our RSRefSeg variants and for other published models\protect\footnotemark[1]. Historic scores are shown in italics, and "--" indicates measurements that are not reported in the cited work.}
\label{tab:combined_training_results}
\resizebox{\textwidth}{!}{%
\begin{tabular}{@{}l|cc|cc|cc|cc@{}}
\toprule
\multirow{2}{*}{\rule{0pt}{2.6ex}\textbf{Model}} & \multicolumn{2}{c|}{\textbf{RefSegRS}} & \multicolumn{2}{c|}{\textbf{RRSIS-D}} & \multicolumn{2}{c|}{\textbf{NWPU-Refer}} & \multicolumn{2}{c}{\textbf{Urban1960SatSeg}} \\
\cmidrule(lr){2-3} \cmidrule(lr){4-5} \cmidrule(lr){6-7} \cmidrule(lr){8-9}
 & \textbf{mIoU} & \textbf{oIoU} & \textbf{mIoU} & \textbf{oIoU} & \textbf{mIoU} & \textbf{oIoU} & \textbf{mIoU} & \textbf{oIoU} \\
\midrule
RMSIN\textsuperscript{1}\cite{liu2024rotated,yang2024large,chen2025rsrefseg} & 59.96\% & 76.81\% & 62.27\% & 76.50\% & 41.75\% & 62.66\% & -- & -- \\
RSRefSeg-B\textsuperscript{1}~\cite{chen2025rsrefseg} & -- & -- & 63.68\% & 76.05\% & -- & -- & -- & -- \\
RSRefSeg-L\textsuperscript{1}~\cite{chen2025rsrefseg} & -- & -- & 64.67\% & 77.24\% & -- & -- & -- & -- \\
MRSNet\textsuperscript{1}~\cite{yang2024large} & -- & -- & -- & -- & 44.86\% & \textbf{63.59\%} & -- & -- \\
Urban1960SatUSM\textsuperscript{1}~\cite{hao2025urban1960satseg} & -- & -- & -- & -- & -- & -- & 68.80\% & -- \\
FIANet\textsuperscript{1}~\cite{lei2025fianet} & 68.67\% & 78.32\% & 64.01\% & 76.91\% & -- & -- & -- & -- \\
SBANet\textsuperscript{1}~\cite{li2025sbanet} & 62.73\% & 79.86\% & 65.52\% & 79.22\% & -- & -- & -- & -- \\
BTDNet\textsuperscript{1}~\cite{zhang2025btdnet} & 67.09\% & 80.57\% & 66.04\% & 79.23\% & -- & -- & -- & -- \\
SegEarth-R1\textsuperscript{1}~\cite{li2025segearth} & 72.45\% & 79.00\% & 66.40\% & 78.01\% & -- & -- & -- & -- \\
RS2-SAM2\textsuperscript{1}~\cite{rong2025rs2sam2} & \textbf{73.90\%} & \textbf{80.87\%} & \textbf{66.72\%} & \textbf{78.99\%} & -- & -- & -- & -- \\
\midrule
\textbf{RSRefSeg-B (ours)} & 24.81\% / \emph{17.54\%} & 40.89\% / \emph{29.41\%} & 64.37\% / \emph{61.16\%} & 76.83\% / \emph{75.44\%} & 39.42\% / \emph{33.15\%} & 59.52\% / \emph{56.56\%} & \textbf{70.65\%} & \textbf{88.86\%} \\
\textbf{RSRefSeg-L (ours)} & 44.52\% / \emph{36.03\%} & 55.74\% / \emph{45.74\%} & 65.37\% / \emph{62.61\%} & 76.33\% / \emph{76.03\%} & \textbf{45.75\%} / \emph{39.11\%} & 62.75\% / \emph{55.29\%} & 69.74\% & 88.73\% \\
\bottomrule
\end{tabular}%
}
\end{table*}
\footnotetext[1]{Published models were trained on individual target datasets without multi-dataset mixing or historic augmentation, as reported in their respective papers. Our models employ unified training across five datasets (Aerial-D, RefSegRS, RRSIS-D, NWPU-Refer, and Urban1960SatSeg) with 20\% historic filter augmentation applied during training.}

\subsection{Ablating LLM Expression Refinement}
\label{subsec:ablation_studies}

To measure how synthetic language affects segmentation quality, we retrained the RSRefSeg model on Aerial-D while isolating the different expression sources. The goal was to contrast the original rule-generated sentences against those produced by the LLM method, and determine which variants provide the strongest supervision signal. We adopt the lighter RSRefSeg-B configuration, with SAM-ViT-Base and LoRA rank $r=16$, so that each run completes quickly while preserving the optimization settings from Section~\ref{subsec:experimental_setup}.

Using this setup, we trained four separate models, each exposed to a distinct slice of Aerial-D: (i) \emph{Rule-based Only} retains the deterministic descriptions produced by the rule-based method; (ii) \emph{LLM Language Variations} relies on rewrites that diversify the wording while preserving the target; (iii) \emph{LLM Visual Variations} selects LLM augmentations that inject alternative visual cues; Finally, (iv) \emph{Combined} combines the three sources. Figure~\ref{fig:llm_enhancement_example} illustrates how the enhanced variants expand the phrasing beyond the rule-based baseline. We evaluate the resulting checkpoints on four test sets without additional tuning: Aerial-D (using the full 405K-expression validation split) and three external datasets, namely RefSegRS, RRSIS-D, and NWPU-Refer, to observe how each expression type supports generalization beyond the training distribution.

Table~\ref{tab:ablation_expression_types} summarizes the obtained results and explicitly reports both the number of \emph{Samples} and \emph{Epochs} used per configuration. Looking across datasets, several consistent patterns emerge. On Aerial-D, the \emph{Combined All} configuration achieves the best accuracy, which is expected given that the validation distribution matches that training mixture. For the three external datasets, different subsets yield the strongest generalization: on RRSIS-D, the \emph{LLM Language Variations} run delivers the highest scores; on NWPU-Refer, emphasizing varied visual cues through \emph{LLM Visual Variations} is most beneficial; on RefSegRS, combining the three sources provides the best results. These outcomes highlight the breadth of ways to phrase referring expressions, and show that leveraging LLMs to introduce targeted variety in training data mixtures can improve cross-dataset generalization.

In order to keep early stopping responsive to each subset, we monitored the  validation loss across the four runs and halted training as soon as it begins to rebound. Because the \emph{Combined} subset is roughly three times larger than the others, its validation loss ticks upward immediately after the second epoch, so we stop that run at two epochs. The three smaller subsets continue improving through the fourth epoch before showing the same rise, allowing four full passes over the data. This schedule means that the language-variation and visual-variation models actually process fewer total samples than the combined run, but still they surpass it on RRSIS-D and NWPU-Refer. This outcome reveals that curated expression subsets can be more sample efficient than the full mixture, which is why the combined model in Section~\ref{subsec:evaluation_results} draws on the LLM Visual Variations split of Aerial-D. Beyond sample efficiency, constraining Aerial-D to that subset keeps the cross-dataset mixture from being dominated by a single corpus, whose referring expression pool would otherwise reach into the millions of training instances.

To further assess the unique challenges introduced by Aerial-D, we trained two additional models. The first model trains exclusively on the three other benchmark datasets, namely RefSegRS, RRSIS-D, and NWPU-Refer, without any Aerial-D supervision. The \emph{Other Datasets} row in Table~\ref{tab:ablation_expression_types} shows that this model achieves strong performance on the original benchmarks, reaching 61.63\% mIoU on RRSIS-D, 49.65\% on RefSegRS, and 44.51\% on NWPU-Refer, demonstrating effective learning from these established corpora. However, when evaluated on Aerial-D, performance drops sharply to only 14.47\% mIoU and 17.77\% oIoU. This substantial degradation underscores the distinct challenges that Aerial-D introduces, including novel semantic categories and instance groups, particularly challenging referring expressions with complex spatial and relational cues from both rule-based generation and LLM enhancement, small-scale instances, and extreme object densities. Conversely, the \emph{LLM Visual Variations + Other Datasets} configuration combines Aerial-D supervision with the three external benchmarks. This combined training delivers improvements on RRSIS-D compared to training without Aerial-D, while showing performance trade-offs on RefSegRS and NWPU-Refer, demonstrating that Aerial-D contributes additional supervision signals that can benefit certain dataset characteristics.

\begin{table*}[t]
\centering
\caption{Results for ablation tests on the expression enhancement component across four datasets.}
\label{tab:ablation_expression_types}
\resizebox{\textwidth}{!}{%
\begin{tabular}{@{}lcc|cc|cc|cc|cc@{}}
\toprule
\multirow{2}{*}{\rule{0pt}{2.6ex}\textbf{Training Configuration}} & \multirow{2}{*}{\rule{0pt}{2.6ex}\textbf{Samples}} & \multirow{2}{*}{\rule{0pt}{2.6ex}\textbf{Epochs}} & \multicolumn{2}{c|}{\textbf{Aerial-D}} & \multicolumn{2}{c|}{\textbf{RefSegRS}} & \multicolumn{2}{c|}{\textbf{RRSIS-D}} & \multicolumn{2}{c}{\textbf{NWPU-Refer}} \\
\cmidrule(lr){4-5} \cmidrule(lr){6-7} \cmidrule(lr){8-9} \cmidrule(lr){10-11}
 & & & \textbf{mIoU} & \textbf{oIoU} & \textbf{mIoU} & \textbf{oIoU} & \textbf{mIoU} & \textbf{oIoU} & \textbf{mIoU} & \textbf{oIoU} \\
\midrule
Rule-based Only & 371K & 4 & 34.57\% & 39.31\% & 3.73\% & 0.55\% & 34.22\% & 36.46\% & 16.78\% & 13.70\% \\
LLM Language Variations & 364K & 4 & 46.45\% & 56.99\% & 5.75\% & 4.99\% & \textbf{41.63\%} & \textbf{42.48\%} & 21.89\% & 16.68\% \\
LLM Visual Variations & 382K & 4 & 46.54\% & 63.02\% & 18.32\% & 8.37\% & 31.78\% & 33.73\% & \textbf{24.68\%} & \textbf{29.22\%} \\
Combined & 1,118K & 2 & \textbf{49.33\%} & \textbf{64.30\%} & \textbf{18.80\%} & \textbf{8.58\%} & 34.07\% & 34.80\% & 24.57\% & 28.27\% \\
\midrule
Other Datasets & 71.6K & 14 & 14.47\% & 17.77\% & \textbf{49.65\%} & \textbf{67.90\%} & 61.63\% & 76.46\% & \textbf{44.51\%} & \textbf{64.22\%} \\
LLM Visual Variations + Other Datasets & 453.6K & 3 & \textbf{45.59\%} & \textbf{62.75\%} & 42.36\% & 56.98\% & \textbf{63.09\%} & \textbf{76.75\%} & 38.18\% & 59.60\% \\
\bottomrule
\end{tabular}%
}
\end{table*}

\subsection{Comparing Gemma3 and OpenAI o3}
\label{subsec:distillation_ablation}

We also attempted to see how different generator choices, inside the LLM enhancement stage, can affect expression quality and overall cost. We compared three options for producing the enhanced expressions: OpenAI's o3, the off-the-shelf Gemma3-12B model, and the distilled Gemma3-Aerial model described in Section~\ref{subsec:llm_expression_generation}. Finally, we include qualitative visualizations to illustrate how RSRefSeg-L responds to historic images and their contemporary counterparts.

All three LLMs are prompted in the same way and used with the same decoding strategy, so that differences stem from the generator rather than the interface. The base Gemma3 model frequently ignores the dual-task schema, hallucinates objects that are not present, and references the guiding bounding boxes inside the expression, which undermines segmentation training. Distillation sharply reduces these errors, producing grounded descriptions that resemble the outputs from OpenAI o3, while remaining accessible on local hardware.

\begin{figure*}[t]
\centering
\begin{minipage}{0.42\textwidth}
\centering
\includegraphics[width=0.7\textwidth]{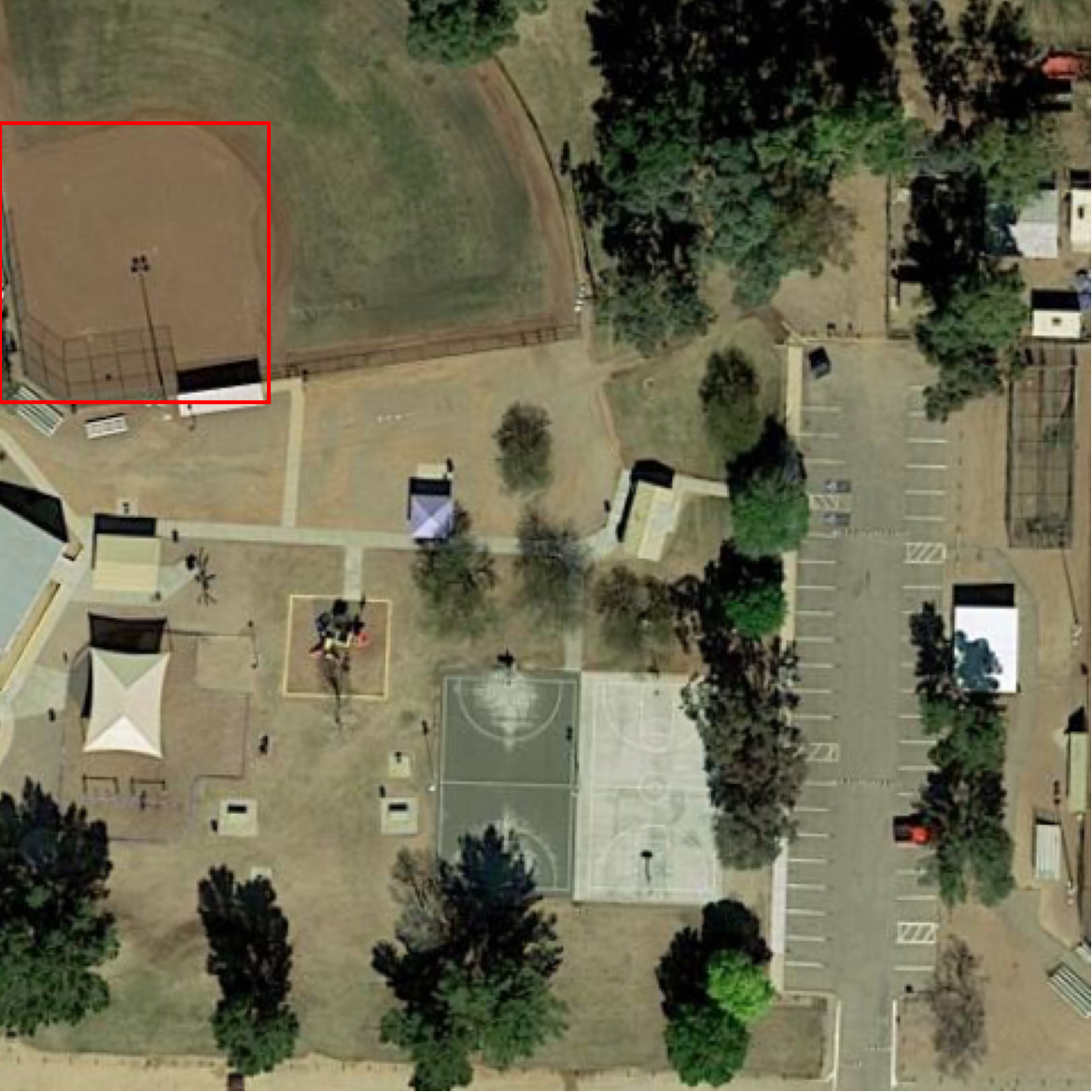}
\end{minipage}%
\begin{minipage}{0.58\textwidth}
\centering
\hspace{-1cm}
\raisebox{-0.2\height}{%
\footnotesize
\begin{tabular}{@{}p{2.5cm}p{6.5cm}@{}}
\toprule
\textbf{Expression Type} & \textbf{Example} \\
\midrule
Original & \emph{the orange baseball diamond in the top-left} \\
\midrule
OpenAI o3 & \emph{the orange baseball diamond with the light pole near home plate in the upper-left region} \\
\midrule
Gemma3 Base & \emph{the bright orange baseball diamond to the left of another similar baseball diamond in the top-left} \\
\midrule
Gemma3-Aerial-12B & \emph{the orange baseball field with a chainlink fence surrounded by grass to the north and trees to the west} \\
\bottomrule
\end{tabular}%
}
\end{minipage}
\caption{Qualitative comparison between OpenAI o3, the base Gemma3 model, and our fine-tuned Gemma3 model on the same aerial scene. Each model receives the same rule-based expression, prompt, and decoding setup, revealing how the rewritten expressions differ under matched conditions.}
\label{fig:distillation_comparison}
\end{figure*}

Table~\ref{tab:cost_comparison} quantifies the practical impact. Running o3 across 259,709 targets would cost roughly \$5.38K, whereas the distilled Gemma3 produces comparable guidance for about \$22. Figure~\ref{fig:distillation_comparison} illustrates how the distilled model mirrors the grounded detail that OpenAI o3 provides, while avoiding the hallucinations seen in the base Gemma3 outputs.

\begin{table}[t]
\centering
\caption{Cost analysis comparing Gemma3 against the OpenAI o3 model for large-scale annotation\protect\footnotemark.}
\label{tab:cost_comparison}
\resizebox{0.8\columnwidth}{!}{%
\begin{tabular}{@{}lcc@{}}
\toprule
\textbf{Model} & \textbf{Cost per request} & \textbf{Cost for 259{,}709 requests} \\
\midrule
OpenAI o3 & \$0.020728 & \$5,382.98 \\
Distilled Gemma3 & \$0.000087 & \$22.80 \\
\midrule
\textbf{Savings} & \textbf{$238\times$ cheaper} & \textbf{\$5,360.18 (99.6\%)} \\
\bottomrule
\end{tabular}%
}
\end{table}
\footnotetext{Cost calculations based on API pricing at the time of making the study-- OpenAI o3 with $2.00$ per million input tokens, and $8.00$ per million output tokens (OpenAI API platform); Gemma3-12B with $0.035$ per million input tokens, and $0.141$ per million output tokens (OpenRouter inference provider). Average tokens per request for OpenAI o3 at $1{,}670.8$ input and $2{,}173.3$ output tokens, and Gemma3 with $1{,}330.0$ input and $284.7$ output tokens. Calculations based on 15 sample requests.}

\subsection{Ablating the Historic Filters}
\label{subsec:historic_ablation}

In order to understand how much the historic image filters, described in Section~\ref{subsec:historic_filters}, contribute to robustness, we repeated the combined training without injecting those filters. This ablation test also used the RSRefSeg-B variant. Models that only encounter clean, contemporary imagery typically falter when historic photographs suddenly introduce monochrome toning or sepia casts.

Table~\ref{tab:historic_ablation_results} summarises three ablation configurations. Each row lists the clean mIoU scores (Orig.), the historic counterpart (Hist.), and the percentage-point delta relative to the full-training baseline presented in Table~\ref{tab:combined_training_results}. The first row removes synthetic historic filters while maintaining Urban1960SatSeg supervision. As expected, performance decreases on the historic variants of RRSIS-D and NWPU-Refer, dropping four and one percentage points, respectively, while Urban1960SatSeg also shows a modest decline. The second row excludes Urban1960SatSeg entirely while retaining the synthetic filters. Here, the Urban1960SatSeg score collapses to below 5\% mIoU, reflecting the substantial distribution shift between Urban1960SatSeg and the remaining datasets. It should be noted that Urban1960SatSeg emphasizes land coverage classes rather than discrete instances, and its imagery exhibits higher brightness and lower resolution characteristics that somewhat diverge from the synthetic historic filters applied to the modern datasets. The RRSIS-D historic score remains unchanged relative to the baseline, demonstrating that synthetic filters alone provide sufficient robustness for this dataset. Interestingly, RefSegRS shows substantial gains on the historic variant in both the first and second rows. These increases likely reflect two compounding factors. First, applying historic filters to 20\% of the training images removes valuable clean data that would otherwise support RefSegRS, which exhibits distribution characteristics distinct from the other datasets. Second, the Urban1960SatSeg supervision may conflict with RefSegRS patterns, and excluding it allows the model to adjust more effectively to the filtered images despite the reduction in overall training data. The third row combines both ablations, removing filters and Urban1960SatSeg data simultaneously, and produces results similar to the first row on RRSIS-D and NWPU-Refer, while maintaining the severe degradation on Urban1960SatSeg. The slightly larger drop on Urban1960SatSeg, seen in this configuration, underscores the compounded effect of eliminating both real historic data and synthetic augmentation.

\begin{table*}[t]
\centering
\caption{Results for ablation tests on the historial image filters. Each column lists the clean score, followed by the historic filter variant shown in italics, with percentage-point deltas relative to Table~\ref{tab:combined_training_results}. The second row removes Urban1960SatSeg supervision while keeping filters. The third row removes both filters and Urban1960SatSeg.}
\label{tab:historic_ablation_results}
\renewcommand{\arraystretch}{1.1}
\resizebox{\textwidth}{!}{%
\begin{tabular}{@{}l|cc|cc|cc|c@{}}
\toprule
\multirow{2}{*}{\rule{0pt}{2.6ex}\textbf{Training Setup}} & \multicolumn{2}{c|}{\textbf{RRSIS-D (mIoU)}} & \multicolumn{2}{c|}{\textbf{NWPU-Refer (mIoU)}} & \multicolumn{2}{c|}{\textbf{RefSegRS (mIoU)}} & \multicolumn{1}{c}{\textbf{Urban1960SatSeg (mIoU)}} \\
\cmidrule(lr){2-3} \cmidrule(lr){4-5} \cmidrule(lr){6-7}
 & \textbf{Orig.} & \textbf{Hist.} & \textbf{Orig.} & \textbf{Hist.} & \textbf{Orig.} & \textbf{Hist.} \\
\midrule
No Filters & 64.29\% & \emph{56.88\%} (\textcolor{red}{-4.28}) & 41.41\% & \emph{32.40\%} (\textcolor{red}{-0.75}) & 44.25\% & \emph{25.47\%} (\textcolor{darkgreen}{+7.93}) & 68.88\% (\textcolor{red}{-1.77}) \\
No Urban1960SatSeg & 64.41\% & \emph{61.16\%} ($\pm$ 0.00) & 40.70\% & \emph{34.89\%} (\textcolor{darkgreen}{+1.74}) & 44.05\% & \emph{34.51\%} (\textcolor{darkgreen}{+16.97}) & 4.87\% (\textcolor{red}{-65.78}) \\
No Filters + No Urban1960SatSeg & 62.75\% & \emph{56.93\%} (\textcolor{red}{-4.23}) & 39.96\% & \emph{30.13\%} (\textcolor{red}{-3.02}) & 43.81\% & \emph{25.61\%} (\textcolor{darkgreen}{+8.07}) & 4.26\% (\textcolor{red}{-66.39}) \\
\bottomrule
\end{tabular}%
}
\renewcommand{\arraystretch}{1}
\end{table*}

Values in parentheses denote percentage-point change relative to the baseline combined model in Table~\ref{tab:combined_training_results}, while italicised values denote historic-filtered validation scores.

\section{Conclusion and Future Work}
\label{sec:conclusion}

This work introduced the Aerial-D dataset together with an end-to-end methodology that converts existing aerial segmentation datasets into referring expression segmentation data. The data processing pipeline features a rule-driven generator that translates segmentation masks into natural language descriptions grounded on absolute positioning, appearance, and relational cues. The rule-based expressions are also refined through LLM rewriting, while keeping costs manageable via a distilled Gemma3 annotator. The resulting corpus enables the training of a generalized RSRefSeg model, establishing reproducible baselines on Aerial-D and remaining competitive with published results on RRSIS-D, NWPU-Refer, RefSegRS, and Urban1960SatSeg. By pairing these evaluations with ablations on expression sources and historic image filters, we demonstrate that Aerial-D delivers a challenging and useful dataset for referring expression segmentation in aerial photos, and highlights the specific ingredients that improve robustness.

Future work can extend this foundation in several promising directions. First, the expression enhancement tool can be applied directly to the native captions supplied with public datasets such as RRSIS-D and NWPU-Refer, enriching their language with additional visual details and linguistic diversity, while preserving the original referring intent, creating a more varied training mixture across the RRSIS ecosystem. Second, achieving state-of-the-art performance on Aerial-D represents a challenging endeavor requiring strong architectures capable of handling dense object distributions, diverse target types, and the complexity of the referring expressions themselves. The interaction between nuanced language and subtle visual cues makes precise grounding particularly difficult, as models must parse intricate spatial descriptions while attending to fine-grained details that distinguish similar objects in crowded scenes. Developing novel architectures, inspired by recent developments~\cite{li2025sam3} and exploring more powerful vision backbones, or incorporating specialized modules for handling small objects and ambiguous spatial relationships could unlock significant performance gains. Third, multilingual variants of these expressions can also be produced while preserving full automation, by pairing high-quality translation models with our recipe—either prompting OpenAI o3 to translate into different languages and training a Gemma3 student to mimic those translations, or seeding the process with dedicated translation models such as Tower Instruct~\cite{tower}. Finally, emerging multimodal systems like Gemini\,2.5~\cite{gemini25} can already output full segmentation masks and bounding boxes. Proper filtering and integration of these results could expand Aerial-D with additional targets derived from the same imagery, which can then be described with our expression-generation stages to unlock richer open-vocabulary supervision.

\section*{Acknowledgments}

This research was supported by the Portuguese Recovery and Resilience Plan through project C645008882-00000055 (i.e., the Center For Responsible AI), and by Fundação para a Ciência e Tecnologia, I.P. (FCT) through the projects with references UID/50021/2025 and UID/PRR/50021/2025.

\bibliographystyle{IEEEtran}
\bibliography{ExtendedAbstract_ref_db}

@inproceedings{zamir2019isaid,
  title        = {{iSAID}: A Large-scale Dataset for Instance Segmentation in Aerial Images},
  author       = {Zamir, Syed Waqas and Arora, Aditya and Gupta, Akshita and Khan, Salman and Sun, Guolei and Khan, Fahad Shahbaz and Zhu, Fan and Shao, Ling and Xia, Gui-Song and Bai, Xiang},
  booktitle    = {Proceedings of the IEEE/CVF Conference on Computer Vision and Pattern Recognition Workshops},
  year         = {2019},
  note         = {CVPR'19 Workshops; arXiv:1905.12886}
}

@inproceedings{xia2018dota,
  title        = {{DOTA}: A Large-Scale Dataset for Object Detection in Aerial Images},
  author       = {Xia, Gui-Song and Bai, Xiang and Ding, Jian and Zhu, Zhen and Belongie, Serge and Luo, Jiebo and Datcu, Mihai and Pelillo, Marcello and Zhang, Liangpei},
  booktitle    = {Proceedings of the IEEE/CVF Conference on Computer Vision and Pattern Recognition},
  year         = {2018},
  publisher    = {IEEE},
  doi          = {10.1109/CVPR.2018.00418}
}

@inproceedings{wang2021loveda,
  title        = {{LoveDA}: A Remote Sensing Land-Cover Dataset for Domain Adaptive Semantic Segmentation},
  author       = {Wang, Junjue and Zheng, Zhuo and Ma, Ailong and Lu, Xiaoyan and Zhong, Yanfei},
  booktitle    = {Proceedings of the Conference on Neural Information Processing Systems Datasets and Benchmarks Track},
  year         = {2021},
}

@inproceedings{ester1996density,
  title        = {A Density-Based Algorithm for Discovering Clusters in Large Spatial Databases with Noise},
  author       = {Ester, Martin and Kriegel, Hans-Peter and Sander, J{\"o}rg and Xu, Xiaowei},
  booktitle    = {Proceedings of the International Conference on Knowledge Discovery and Data Mining},
  year         = {1996},
  publisher    = {AAAI Press}
}

@article{yuan2023rrsis,
  title        = {{RRSIS}: Referring Remote Sensing Image Segmentation},
  author       = {Yuan, Zhenghang and Mou, Lichao and Hua, Yuansheng and Zhu, Xiao Xiang},
  journal      = {IEEE Transactions on Geoscience and Remote Sensing},
  volume       = {62},
  year         = {2024},
  publisher    = {IEEE},
  doi          = {10.1109/TGRS.2024.3369720}
}

@inproceedings{liu2024rotated,
  title        = {Rotated Multi-Scale Interaction Network for Referring Remote Sensing Image Segmentation},
  author       = {Liu, Sihan and Ma, Yiwei and Zhang, Xiaoqing and Wang, Haowei and Ji, Jiayi and Sun, Xiaoshuai and Ji, Rongrong},
  booktitle    = {Proceedings of the IEEE/CVF Conference on Computer Vision and Pattern Recognition},
  year         = {2024},
  publisher    = {IEEE},
  doi          = {10.1109/CVPR52733.2024.02517}
}

@article{li2020dior,
  title        = {Object Detection in Optical Remote Sensing Images: A Survey and a New Benchmark},
  author       = {Li, Ke and Wan, Guorun and Cheng, Gong and Meng, Lichao and Shi, Jian},
  journal      = {ISPRS Journal of Photogrammetry and Remote Sensing},
  volume       = {165},
  year         = {2020},
  publisher    = {Elsevier},
  doi          = {10.1016/j.isprsjprs.2020.04.001}
}

@article{yang2024large,
  title        = {A Large-Scale Referring Remote Sensing Image Segmentation Dataset and Benchmark},
  author       = {Yang, Zhigang and Yao, Huiguang and Tian, Linmao and Zhao, Xuezhi and Li, Qiang and Wang, Qi},
  journal      = {arXiv preprint arXiv:2506.03583},
  year         = {2025}
}

@article{hao2025urban1960satseg,
  title        = {{Urban1960SatSeg}: Unsupervised Semantic Segmentation of Mid-20th Century Urban Landscapes with Satellite Imageries},
  author       = {Hao, Tianxiang and Zhang, Lixian and Zhang, Yingjia and Chen, Mengxuan and Zhang, Jinxiao and Fu, Haohuan},
  journal      = {arXiv preprint arXiv:2506.09476},
  year         = {2025}
}

@article{chen2025rsrefseg,
  title        = {{RSRefSeg}: Referring Remote Sensing Image Segmentation with Foundation Models},
  author       = {Chen, Keyan and Zhang, Jiafan and Liu, Chenyang and Zou, Zhengxia and Shi, Zhenwei},
  journal      = {arXiv preprint arXiv:2501.06809},
  year         = {2025}
}

@article{sam,
  title        = {Segment Anything},
  author       = {Kirillov, Alexander and Mintun, Eric and Ravi, Nikhila and Mao, Hanzi and Rolland, Chloe and Gustafson, Laura and Xiao, Tete and Whitehead, Spencer and Berg, Alexander C. and Lo, Wan-Yen and Doll{\'a}r, Piotr and Girshick, Ross},
  journal      = {arXiv preprint arXiv:2304.02643},
  year         = {2023}
}

@inproceedings{clip,
  title        = {Learning Transferable Visual Models From Natural Language Supervision},
  author       = {Radford, Alec and Kim, Jong Wook and Hallacy, Chris and Ramesh, Aditya and Goh, Gabriel and Agarwal, Sandhini and Sastry, Girish and Askell, Amanda and Mishkin, Pamela and Clark, Jack and Krueger, Gretchen and Sutskever, Ilya},
  booktitle    = {Proceedings of the International Conference on Machine Learning},
  volume       = {139},
  year         = {2021},
  publisher    = {PMLR}
}

@inproceedings{siglip,
  title        = {Sigmoid Loss for Language Image Pre-Training},
  author       = {Zhai, Xiaohua and Mustafa, Basil and Kolesnikov, Alexander and Beyer, Lucas},
  booktitle    = {Proceedings of the IEEE/CVF International Conference on Computer Vision},
  year         = {2023},
  publisher    = {IEEE},
  doi          = {10.1109/ICCV51070.2023.01100}
}

@article{siglip2,
  title        = {{SigLIP 2}: Multilingual Vision-Language Encoders with Improved Semantic Understanding, Localization, and Dense Features},
  author       = {Tschannen, Michael and Gritsenko, Alexey and Wang, Xiao and Naeem, Muhammad Ferjad and Alabdulmohsin, Ibrahim and Parthasarathy, Nikhil and Evans, Talfan and Beyer, Lucas and Xia, Ye and Mustafa, Basil and H{\'e}naff, Olivier and Harmsen, Jeremiah and Steiner, Andreas and Zhai, Xiaohua},
  journal      = {arXiv preprint arXiv:2502.14786},
  year         = {2025}
}

@inproceedings{lora,
  title        = {{LoRA}: Low-Rank Adaptation of Large Language Models},
  author       = {Hu, Edward J. and Shen, Yelong and Wallis, Phillip and Allen-Zhu, Zeyuan and Li, Yuanzhi and Wang, Lu and Chen, Weizhu},
  booktitle    = {Proceedings of the International Conference on Learning Representations},
  year         = {2022}
}

@inproceedings{adamw,
  title        = {Decoupled Weight Decay Regularization},
  author       = {Loshchilov, Ilya and Hutter, Frank},
  booktitle    = {Proceedings of the International Conference on Learning Representations},
  year         = {2019}
}

@inproceedings{qlora,
  title        = {{QLoRA}: Efficient Finetuning of Quantized {LLMs}},
  author       = {Dettmers, Tim and Pagnoni, Artidoro and Holtzman, Ari and Zettlemoyer, Luke},
  booktitle    = {Proceedings of the Conference on Neural Information Processing Systems},
  year         = {2023}
}

@misc{gemini25,
  title        = {Gemini 2.5: Multimodal Model Overview},
  author       = {{Google DeepMind}},
  year         = {2025},
  howpublished = {Product and research overview},
  url          = {https://deepmind.google/technologies/gemini/},
  note         = {Accessed 2025-01-15}
}

@misc{marvin2023chai,
  author       = {Marvin, Benjamin and Zambanini, Sebastian and Philipp, Patrick},
  title        = {Craters in Historical Aerial Images (CHAI) Dataset},
  howpublished = {Zenodo},
  year         = {2023},
  month        = nov,
  doi          = {10.5281/zenodo.10068633}
}

@inproceedings{swin,
  title        = {Swin Transformer: Hierarchical Vision Transformer using Shifted Windows},
  author       = {Liu, Ze and Lin, Yutong and Cao, Yue and Hu, Han and Wei, Yixuan and Zhang, Zheng and Lin, Stephen and Guo, Baining},
  booktitle    = {Proceedings of the IEEE/CVF International Conference on Computer Vision},
  year         = {2021},
  publisher    = {IEEE},
  doi          = {10.1109/ICCV48922.2021.00986}
}

@inproceedings{bert,
  title        = {{BERT}: Pre-training of Deep Bidirectional Transformers for Language Understanding},
  author       = {Devlin, Jacob and Chang, Ming-Wei and Lee, Kenton and Toutanova, Kristina},
  booktitle    = {Proceedings of the Conference of the North American Chapter of the Association for Computational Linguistics: Human Language Technologies},
  year         = {2019},
  publisher    = {Association for Computational Linguistics},
  doi          = {10.18653/v1/N19-1423}
}

@inproceedings{vit,
  title        = {An Image is Worth 16x16 Words: Transformers for Image Recognition at Scale},
  author       = {Dosovitskiy, Alexey and Beyer, Lucas and Kolesnikov, Alexander and Weissenborn, Dirk and Zhai, Xiaohua and Unterthiner, Thomas and Dehghani, Mostafa and Minderer, Matthias and Heigold, Georg and Gelly, Sylvain and Uszkoreit, Jakob and Houlsby, Neil},
  booktitle    = {Proceedings of the International Conference on Learning Representations},
  year         = {2021},
}

@inproceedings{pytorch,
  title        = {PyTorch: An Imperative Style, High-Performance Deep Learning Library},
  author       = {Paszke, Adam and Gross, Sam and Massa, Francisco and Lerer, Adam and Bradbury, James and Chanan, Gregory and Killeen, Trevor and Lin, Zeming and Gimelshein, Natalia and Antiga, Luca and Desmaison, Alban and K{\"o}pf, Andreas and Yang, Edward and DeVito, Zachary and Raison, Martin and Tejani, Alykhan and Chilamkurthy, Sasank and Steiner, Benoit and Fang, Lu and Bai, Junjie and Chintala, Soumith},
  booktitle    = {Proceedings of the Conference on Neural Information Processing Systems},
  year         = {2019},
}

@inproceedings{mixedprecision,
  title        = {Mixed Precision Training},
  author       = {Micikevicius, Paulius and Narang, Sharan and Alben, Jonah and Diamos, Gregory and Elsen, Erich and Garcia, David and Ginsburg, Boris and Houston, Michael and Kuchaiev, Oleksii and Venkatesh, Ganesh and Wu, Hao},
  booktitle    = {Proceedings of the International Conference on Learning Representations},
  year         = {2018},
}

@inproceedings{kazemzadeh2014referit,
  title        = {{ReferItGame}: Referring to Objects in Photographs of Natural Scenes},
  author       = {Kazemzadeh, Sahar and Ordonez, Vicente and Matten, Mark and Berg, Tamara},
  booktitle    = {Proceedings of the Conference on Empirical Methods in Natural Language Processing},
  year         = {2014},
  publisher    = {Association for Computational Linguistics},
  doi          = {10.3115/v1/D14-1086}
}

@inproceedings{yu2016modeling,
  title        = {Modeling Context in Referring Expressions},
  author       = {Yu, Licheng and Poirson, Patrick and Yang, Shan and Berg, Alexander C. and Berg, Tamara L.},
  booktitle    = {Proceedings of the European Conference on Computer Vision},
  year         = {2016},
  publisher    = {Springer International Publishing},
  doi          = {10.1007/978-3-319-46475-6_5}
}

@inproceedings{hu2016segmentation,
  title        = {Segmentation from Natural Language Expressions},
  author       = {Hu, Ronghang and Rohrbach, Marcus and Darrell, Trevor},
  booktitle    = {Proceedings of the European Conference on Computer Vision},
  year         = {2016},
  publisher    = {Springer International Publishing},
  doi          = {10.1007/978-3-319-46448-0_7}
}

@article{tower,
  title        = {Tower: An Open Multilingual Large Language Model for Translation-Related Tasks},
  author       = {Alves, Duarte and Guerreiro, Nuno M. and Alves, Jo{\~a}o and Pombal, Jos{\'e} and Rei, Ricardo and Fernandes, Jos{\'e} G. C. and Farinhas, Ant{\'o}nio and Coheur, Lu{\'\i}sa and Martins, Andr{\'e} F. T.},
  journal      = {arXiv preprint arXiv:2402.17733},
  year         = {2024}
}

@article{everingham2010pascal,
  title        = {The {P}ascal {V}isual {O}bject {C}lasses ({VOC}) Challenge},
  author       = {Everingham, Mark and Van Gool, Luc and Williams, Christopher K. I. and Winn, John and Zisserman, Andrew},
  journal      = {International Journal of Computer Vision},
  volume       = {88},
  number       = {2},
  year         = {2010},
  publisher    = {Springer},
  doi          = {10.1007/s11263-009-0275-4}
}

@article{hinton2015distilling,
  title        = {Distilling the Knowledge in a Neural Network},
  author       = {Hinton, Geoffrey and Vinyals, Oriol and Dean, Jeff},
  journal      = {arXiv preprint arXiv:1503.02531},
  year         = {2015},
}

@misc{melbourne1945,
  title        = {Melbourne 1945 --- Historical Aerial Imagery Explorer},
  howpublished = {City of Melbourne, \url{https://1945.melbourne/}},
  note         = {Accessed 2025-01-15}
}

@article{HammerUr2019,
  author       = {Hammer, Emily and Ur, Jason},
  title        = {Near Eastern Landscapes and Declassified {U}-2 Aerial Imagery},
  journal      = {Advances in Archaeological Practice},
  volume       = {7},
  number       = {2},
  year         = {2019},
  doi          = {10.1017/aap.2018.38}
}

@article{lei2025fianet,
  title        = {Exploring Fine-Grained Image-Text Alignment for Referring Remote Sensing Image Segmentation},
  author       = {Lei, Sen and Xiao, Xinyu and Zhang, Tianlin and Li, Heng-Chao and Shi, Zhenwei and Zhu, Qing},
  journal      = {IEEE Transactions on Geoscience and Remote Sensing},
  volume       = {63},
  year         = {2025},
  publisher    = {IEEE}
}

@article{li2025sbanet,
  title        = {Scale-wise Bidirectional Alignment Network for Referring Remote Sensing Image Segmentation},
  author       = {Li, Kun and Vosselman, George and Yang, Michael Ying},
  journal      = {ISPRS Journal of Photogrammetry and Remote Sensing},
  volume       = {226},
  year         = {2025},
  publisher    = {Elsevier}
}

@article{zhang2025btdnet,
  title        = {Referring Remote Sensing Image Segmentation via Bidirectional Alignment Guided Joint Prediction},
  author       = {Zhang, Tianxiang and Wen, Zhaokun and Kong, Bo and Liu, Kecheng and Zhang, Yisi and Zhuang, Peixian and Li, Jiangyun},
  journal      = {arXiv preprint arXiv:2502.08486},
  year         = {2025}
}

@article{li2025segearth,
  title        = {Seg{E}arth-{R}1: Geospatial Pixel Reasoning via Large Language Model},
  author       = {Li, Kaiyu and Xin, Zepeng and Pang, Li and Pang, Chao and Deng, Yupeng and Yao, Jing and Xia, Guisong and Meng, Deyu and Wang, Zhi and Cao, Xiangyong},
  journal      = {arXiv preprint arXiv:2504.09644},
  year         = {2025}
}

@article{rong2025rs2sam2,
  title        = {RS2-{SAM}2: Customized SAM2 for Referring Remote Sensing Image Segmentation},
  author       = {Rong, Fu and Lan, Meng and Zhang, Qian and Zhang, Lefei},
  journal      = {arXiv preprint arXiv:2503.07266},
  year         = {2025}
}

@inproceedings{sertel2023historic,
  title        = {Deep Learning-Based Land Use Land Cover Segmentation of Historical Aerial Images},
  author       = {Sertel, Elif and Avci, Cengiz and Kabadayi, Mustafa Erdem},
  booktitle    = {Proceedings of the IEEE International Geoscience and Remote Sensing Symposium},
  year         = {2023},
  doi          = {10.1109/IGARSS52108.2023.10281819}
}

@article{gemma_3_2025,
  title        = {Gemma 3 Technical Report},
  author       = {{Gemma Team}},
  journal      = {arXiv preprint arXiv:2503.19786},
  year         = {2025}
}

@article{li2025sam3,
  title        = {{SAM3-I}: Segment Anything with Instructions},
  author       = {Li, Jingjing and Feng, Yue and Guo, Yuchen and Huang, Jincai and Piao, Yongri and Bi, Qi and Zhang, Miao and Zhao, Xiaoqi and Chen, Qiang and Zou, Shihao and others},
  journal      = {arXiv preprint arXiv:2512.04585},
  year         = {2025}
}

\begin{IEEEbiography}[{\includegraphics[width=1in,height=1.25in,clip,keepaspectratio]{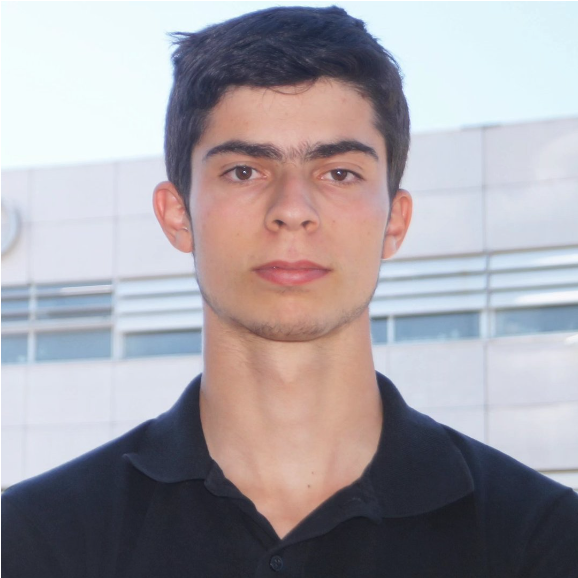}}]{Luís Marnoto}
received the Bachelor's degree in electronics engineering from Instituto Superior Técnico, University of Lisbon, Lisbon, Portugal, in 2023, and the Master's degree in electrical and computer engineering from the same institution in 2025, developed in collaboration with the Human Language Technologies Laboratory, INESC-ID, Lisbon. His research interests include computer vision and multimodal deep learning.
\end{IEEEbiography}

\begin{IEEEbiography}[{\includegraphics[width=1in,height=1.25in,clip,keepaspectratio]{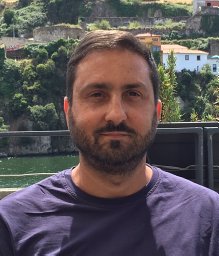}}]{Alexandre Bernardino}
(Senior Member, IEEE) received the Ph.D. degree in 2004. He is a Tenured Associate Professor and a Senior Researcher with the Institute for Systems and Robotics, IST, Lisbon University, Lisbon, Portugal. He published more than 300 research papers and participated in more than 20 national and international research projects, being the Principal Investigator in five of them. His main research interests include the application of computer vision, machine learning, cognitive science, and control theory to advanced robotics and automation systems.
\end{IEEEbiography}

\begin{IEEEbiography}[{\includegraphics[width=1in,height=1.25in,clip,keepaspectratio]{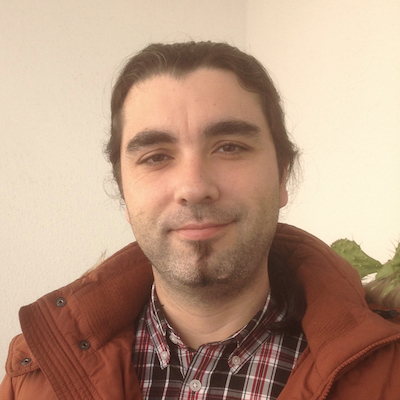}}]{Bruno Martins}
(Senior Member, IEEE) was born in Lisbon, Portugal, in 1979. He received the Ph.D. degree in computer science from the University of Lisbon, in 2009. He is currently an Associate Professor with the Instituto Superior Técnico, University of Lisbon, a Researcher with the Human Language Technologies Laboratory, INESC-ID, and a member of the Lisbon ELLIS Unit (LUMLIS). He works on problems related to the general areas of information retrieval, text mining, multimodal machine learning, and the geographical information sciences. He has been involved in several research projects related to geospatial aspects in information access and retrieval, and he has accumulated a significant expertise in addressing challenges at the intersection of information retrieval and the geographical information sciences.
\end{IEEEbiography}

\end{document}